\pdfoutput=1
\documentclass[11pt]{article}

\usepackage{emnlp2021}
\usepackage{times}
\usepackage{hyperref}
\usepackage{latexsym}
\usepackage{graphicx}
\usepackage{amsfonts,amsmath,mathtools,bm}
\usepackage{booktabs}
\usepackage{float}
\usepackage[T1]{fontenc}
\usepackage[utf8]{inputenc}

\usepackage{microtype}

\title{The Emergence of the Shape Bias Results from Communicative Efficiency}

\author{Eva Portelance\thanks{\enspace Corresponding author.} \\
  Stanford University\thanks{\enspace This project was started during an internship at Microsoft Research Montreal.} \\
  \texttt{portelan@stanford.edu} \\\And
  Michael C. Frank \\
  Stanford University \\
  \texttt{mcfrank@stanford.edu} \\\AND
  Dan Jurafsky\\
  Stanford University \\
  \texttt{jurafsky@stanford.edu} \\\And
  Alessandro Sordoni \\
  Microsoft Research Montreal \\
  \texttt{alsordon@microsoft.com}
  \\\AND
  Romain Laroche \\
  Microsoft Research Montreal \\
  \texttt{romain.laroche@microsoft.com}
  }

\begin{document}
\maketitle
\begin{abstract}
By the age of two, children tend to assume that new word categories are based on objects’ shape, rather than their color or texture; this assumption is called the shape bias. They are thought to learn this bias by observing that their caregiver's language is biased towards shape based categories. This presents a chicken and egg problem: if the shape bias must be present in the language in order for children to learn it, how did it arise in language in the first place? In this paper, we propose that communicative efficiency explains both how the shape bias emerged and why it persists across generations. We model this process with neural emergent language agents that learn to communicate about raw pixelated images. First, we show that the shape bias emerges as a result of efficient communication strategies employed by agents. Second, we show that pressure brought on by communicative need is also necessary for it to persist across generations; simply having a shape bias in an agent’s input language is insufficient. These results suggest that, over and above the operation of other learning strategies, the shape bias in human learners may emerge and be sustained by communicative pressures.
\end{abstract}

\section{Introduction} %Page 1
%\footnote{The problem of referential inscrutability: If a native speaker of an unknown language says `Gavagai!' while seeing a rabbit, how do we determine what this means? It could mean `rabbit', but it could also refer to a more general word like `animal', `white thing', some specific part of the rabbit, or not even refer to the rabbit at all, maybe it could refer to the animal's action, like running away.}

\begin{figure}[t]
\vskip 0.2in
\begin{center}
\centerline{\includegraphics[scale=0.7]{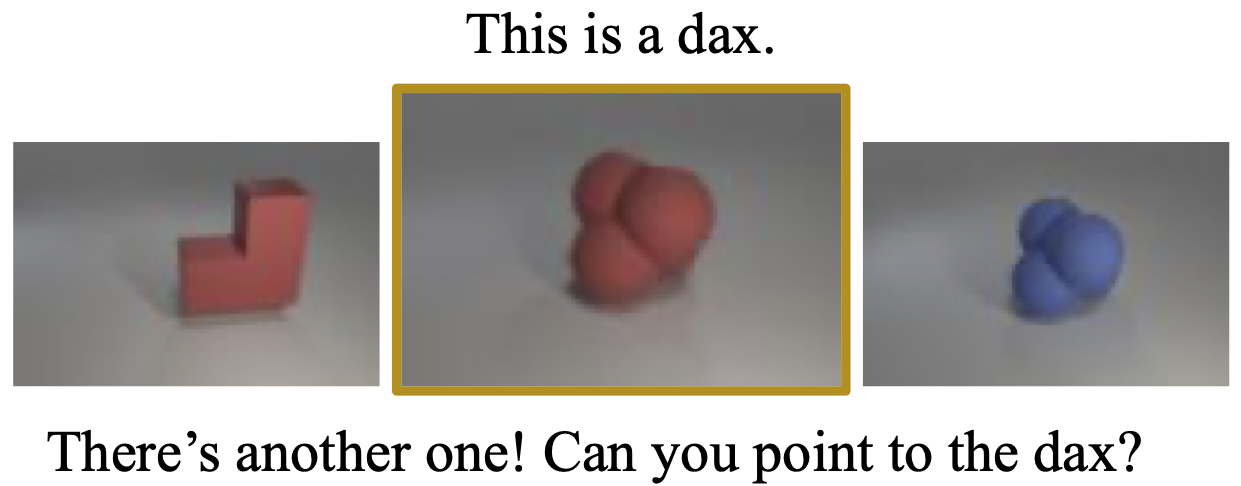}}
\caption{An example experimental trial on novel word learning. Humans typically choose the right picture.}\label{fig:top}
\end{center}
\vskip -0.2in
\end{figure}

Learning new words involves inferring what the referents are in addition to learning a mapping from these inferred referents to linguistic conventions. This is a difficult problem, famously illustrated by \citeauthor{quine1960word}'s (\citeyear{quine1960word}) problem of referential inscrutability. Yet, human children manage just fine. A common explanation is that they must be using additional sources of information to learn new words. One such source is inductive word learning biases, such as the \textit{shape bias}, which appears in children around two years of age \cite{heibeck1987word, landau1988importance}. This bias can be observed when children and adults are presented with a novel object and word. They will often assume that this word refers to a class categorized by the shape of the novel object. Thus, when they are presented with more exemplars they will tend to classify those that share the same shape in the same lexical category, rather than those which share the same color or texture, (Figure \ref{fig:top}).

In this paper, we endeavor to answer two questions: how did the shape bias emerge in our lexical categorization systems? and why does it persist across generations, languages, and cultures? Using neural emergent communication agents as our model of linguistic convention formation, we show how efficient communication strategies can explain both the emergence and persistence of this bias.

Languages have been shown to follow principles of efficient communication for naming systems in a variety of semantic domains, including kinship terms \cite{kemp2012kinship} and color terms \cite{zaslavsky2019color}. A given naming system tends to strike a perfect balance between its overall informativity and simplicity as a function of speakers' need to distinguish referents in that system \cite{kemp2018semantic}. The effects of efficient communication strategies in one semantic domain on other related domains have yet to be considered. In this paper, we demonstrate how efficient communication strategies in one semantic domain -- shape -- can affect the overall lexicon and lead to categorical biases, like the shape bias.

\paragraph{Contributions}
We propose an explanation for the existence and persistence of the shape bias, finding that both follow from principles of efficient communication;
We develop a formal way to model the emergence of lexical categories using neural emergent communication agents\footnote{All the code and data used for this paper are available at \href{https://github.com/evaportelance/emergent-shape-bias}{github.com/evaportelance/emergent-shape-bias.}}; We extend the communicative efficiency paradigm to learning contexts and consider its effects on the overall lexicon.

\section{Previous Work} % Page 2

\subsection{The shape bias in children}

The shape bias is one of many inductive biases that help children learn new words \citep{markman1990constraints}. It was initially studied by \citet{landau1988importance}. They found that the strength of the bias depended on the age of the participants and task, generally appearing around two years old and growing stronger over time. Its acquisition has been attributed to two possible learning processes: associative word learning \cite[also known as attentional learning]{samuelson1999early, samuelson2002statistical, regier2003emergent, regier2005emergence, smith2006attentional}  and conceptual learning \cite{diesendruck2003specific, booth2005conceptual, kemp2007learning}.

Associative learning accounts argue that children observe statistical regularities while learning a language's lexical categories and generalize higher order lexical categorical biases -- for example, by observing that all items in categories `bowl', `cup', or `spoon' share a shape, one might hypothesise that novel lexical categories will also share this property. Conceptual learning accounts argue that the shape bias is a conceptual bias learnt by generalizing categorical features of different kinds, proposing a more abstract view of this inductive bias, that goes beyond lexical categories specifically. Importantly, both of these approaches rely on the existence of bias towards shape in the lexical or conceptual categories of the caregivers a child is exposed to in order for the shape bias to be learnt. Once learnt, this same bias can be used to form new categories, creating a chicken-and-egg problem: the shape bias helps us learn new words and language helps us learn the shape bias.

\subsection{The shape bias in neural networks}

Early connectionist models of associative learning demonstrated how it was possible to learn the shape bias via statistical regularities in lexical categories \cite{samuelson2002statistical, regier2003emergent, regier2005emergence}. As image recognition models developed, the shape bias peaked interest as a possible behavioral benchmark for comparing vision model recognition behavior to human recognition \cite{ritter2017cognitive, hosseini2018assessing, geirhos2018imagenettrained, tuli2021convolutional, bhojanapalli2021understanding}. However, using this bias as a general behavioral benchmark ignores much of the work showing that it is contextually modulated and task dependant in humans; it is not a general purpose perceptual bias -- i.e. we don't necessarily have a shape bias when categorizing natural kinds like foods or substances \cite{imai1997cross, soja1991ontological, yoshida2003shifting, diesendruck2003specific, booth2005conceptual, cimpian2005absence}. With the advent of situated agents and emergent language agents, the use of neural networks to study the learning process for the shape bias reemerged, reproducing similar results to earlier connectionist models supporting associative learning \cite{hill2019understanding} and demonstrating how existing perceptual biases can influence lexical category formation \cite{ohmerand2021}.

\subsection{Communicative efficiency}

Communicative efficiency is an information theoretic approach for explaining both semantic and syntactic linguistic typological observations (for review see \citet{kirby2015compression, kemp2018semantic, gibson2019efficiency}). Here, we focus on semantic domains. Languages must be informative, which means that they must allow a speaker to accurately convey intended meaning to a listener. There is a `communicative cost' associated with \textit{information loss}. Languages must also be simple since we have a limited amount of cognitive resources which must be properly utilized. There is a `cognitive cost' associated with language with higher \textit{complexity}. A language is considered to respect communicative efficiency if it follows the optimal trade-off between informativity and complexity; natural languages have been found to do so for a number of structured semantic domains \cite{gibson2017color, zaslavsky2019color, kemp2012kinship, xu2014numeral}. This trade-off is obtained because both informativity and complexity are modulated by the likelihood of needing to communicate a given semantic distinction, which we refer to as \textit{communicative need}.

Neural emergent communication agents have been shown to learn efficient linguistic mappings from structured semantic space, specifically for color naming systems \cite{Chaabounie2021communicating}.

\subsection{Neural emergent communication agents}

Neural emergent communication agents have been used to study the emergence of structural linguistic properties \cite{andreas-klein-2016-reasoning, lazaridou2017multi, havrylov2017emergence, kottur-etal-2017-natural}. They have also been used to model the emergence of the mutual exclusivity bias -- another child word learning bias \cite{ohmer2020reinforcement}.

Initially, agents performed communication tasks that used structured input, where optimal partition were given by symbolic inputs or by using encodings from pretrained supervised vision models. These tasks only required that agents develop a coherent mapping between these features and their messages. Tasks that require simultaneously learning visual and linguistic representations -- by starting from unstructured input that does not explicitly define the meanings agents must use, like pixelated images -- have been found to be very difficult; agents often failed to learn any meaningful representations \cite{lazaridou2018emergence, bouchacourtbaroni2018agents}. \citet{steels2005coordinating} and \citet{Chaabounie2021communicating} have looked at the emergence of lexical categories using structured input, but to the best of our knowledge it has yet to be done using unstructured input.

\section{A Model of Lexical Category Learning}

In order to model the problem of language learning we first need a task, agent architecture, and learning procedure that allow us to simultaneously learn visual and linguistic representations. Then, we can ask how visual and lexical categorization systems biased towards shape can both (1) arise in a language, and furthermore, (2) persist across generations of language learners.

\subsection{Data and task}

\begin{figure*}[t]
\vskip 0.2in
\begin{center}
\centerline{\includegraphics[scale=0.55]{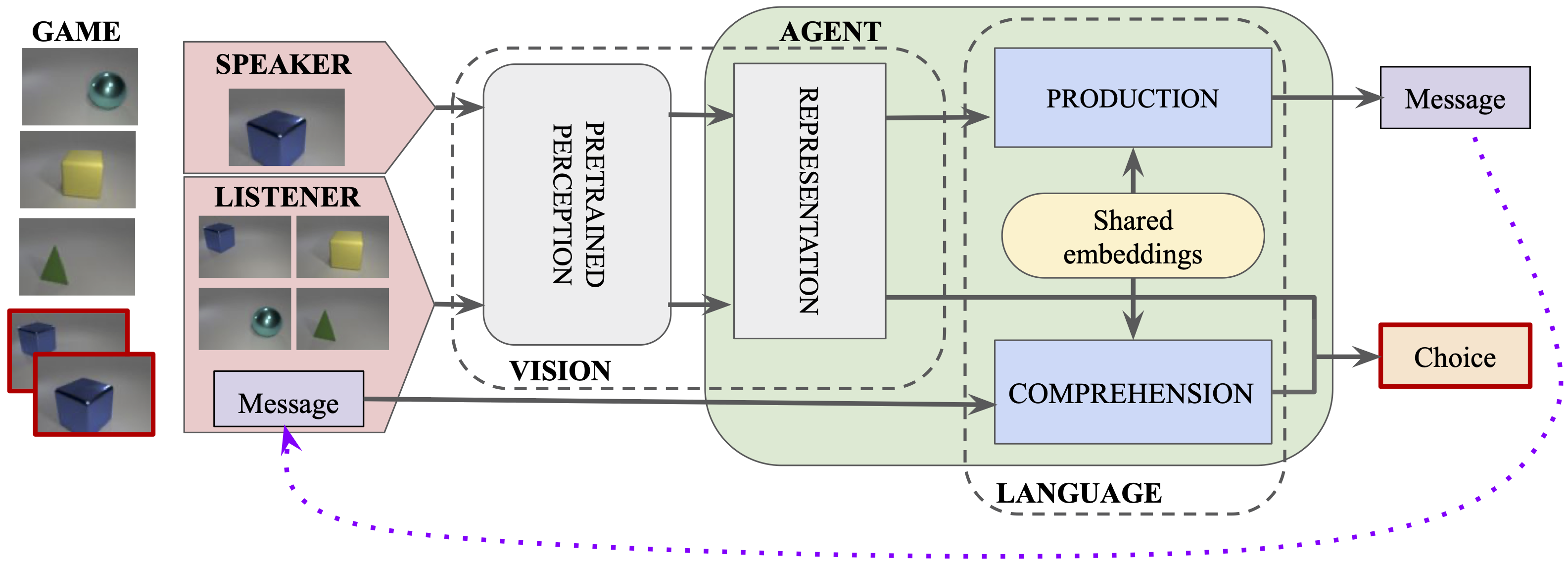}}
\caption{Task and agent architecture design. The \textit{AGENT} box contains modules that update during learning. \label{fig:1}}
\end{center}
\vskip -0.2in
\end{figure*}
We generated a dataset of single object images based on the CLEVR Dataset generator \cite{johnson2017clevr}. Objects can vary in color, shape, material and size. In addition to the existing CLEVR object features, we added 7 additional shapes, for a total of 10 possible shapes (cube, sphere, cylinder, cone, tetrahedron, torus, icosahedron, 6-sided symmetric cross, 3D L-shape, tetrahedral sphere form), 8 colors (gray, red, blue, green, brown, purple, cyan, yellow), 2 materials (metal, rubber), and 2 sizes (small, large), totalling 320 possible combinations as objects.\footnote{Refer to Appendix \ref{app:data} for additional exemplary images.} We randomly varied object and lighting position on a 3D gray background to generate 33,000 images, 22,000 for train set and 11,000 for test set. All objects appear in both test and train but the images and perspectives of these objects vary across sets.

We use the images in our train and test sets to design communication games. Agents play a variation on the classic Lewis signalling game \cite{lewis1969convention} designed to study how communicative partners converge on a system of linguistic conventions referring to these objects.\footnote{Similar variations on Lewis signalling games have been used to study the dynamics of category formation \cite{jager2007evolution, o2014evolving}.} The game is played by a speaker (sender) and a listener (receiver). Speaker S is given an image of the target object $i$ and produces a message $\bm{m} = [m_1,...,m_{n}]$ which is then transmitted to the listener L. Using this message, the listener must then try to correctly identify this target object from a set of images $I$ which contains a different perspective of the target object and additional distractor images.\footnote{When given identical target images agents tend to refer to the position of object in the images rather than its color or shape because coordinates are in continuous space and thus more distinctive, while color and shape are not. We avoid this problem by having agents see different perspectives of the same object.} Thus, agents are maximally cooperative with one another. The images for each game are sampled from the previously described dataset; an exemplary game is available in Figure \ref{fig:1}. In the following experiments, each game consists of one target object and three distractors.

\subsection{Agent architecture}

We describe how the vision and language components of each agent model interact (see Figure \ref{fig:1})\footnote{Refer to Appendix \ref{app:hyp} for full implementation details.}.

\paragraph{Vision}

Training agents directly on pixelated images as input is very difficult: often they are unable to learn a meaningful communication system \cite{lazaridou2018emergence, bouchacourtbaroni2018agents}. We offer a solution to this problem by taking inspiration from humans. We separate image processing into two steps: (1) image perception, and (2) image representation.

For humans, perception is dependent on physiological constraints and perceptive predispositions which are independent from language -- for example newborns' predispositions towards detecting faces \cite{simion2015face} -- while representation is intertwined with language. The distinctive features we choose to describe in an image are dependent on the tools we have to communicate them. Thus, in our agents, we adopt the approach that the perception of an image is independent of how the agents choose to represent and describe it.

For perception, we use a fixed pretrained model. Crucially, this model should not be trained using supervision to recognize specific features. We choose to use AMDIM \cite{bachman2019amdim}, a self-supervised learning model for image encodings. We pretrained this model on our images using its default hyperparameter settings.\footnote{It should be noted that this model does not have `innate' perceptual shape bias; though it may encode shape features, as we will see in experiment 1, it in fact has a tendency towards primarily encoding more localized features like color.}

As for representation, each agent is equipped with their own representation encoder that updates during learning. For each image, the encoder takes the output of the perception module and selectively reduces it to a smaller feature vector before passing it to a language module.\footnote{For additional motivation for the use of this dual vision module approach over previous approaches see Appendix \ref{app:learn}.}

\paragraph{Language}

When the agent is assigned the speaker role, they receive an image of the target object $i$ and must produce a descriptive message $\bm{m}$ for the listener. This image goes through the vision modules and the output of the representation encoder acts as the initial hidden state of the production module, a single layered LSTM. The LSTM sequentially generates the speaker's message. At each state we sample the next character from a categorical distribution over the vocabulary. This character is then used as the input of the subsequent state to generate the next character. The cycle continues until an end-of-string character is produced or the max message length is reached. So for Speaker S, we say that $p_{\textsc{s}}(m_n) = p_{\textsc{s}}(\cdot|m_1, ..., m_{n-1},i)$ for each character $m_n$ in a message $\bm{m}$. The probability of the full message given the image of the target $i$ is $p_{\textsc{s}}(\bm{m}|i) = \prod_n p_{\textsc{s}}(m_n)$.

If the agent is the listener, the message $\bm{m}$ generated by the speaker is used as the input of the comprehension module, here again an LSTM. The final state of the LSTM is combined by a dot product with the output of the representation encoder for each image in the set of images $I$ of a given game. The listener receives a different image of the target object as well as 3 distractor images. The combined message and image interpretation vectors are then used to predict which image contains the target object. The listener's guess $i'$ is sampled from the predicted target distribution over the images in $I$: $i' \sim p_{\textsc{l}}(\cdot|\bm{m},I)$.

Our agents are symmetric \cite{cao2018emergent, bouchacourt-baroni-2019-miss, harding-graesser-etal-2019-emergent}. This means they can alternate between being speaker or listener. This approach allows us to use a shared set of embeddings for the production and comprehension modules within each individual agent in order to represent messages. Additionally, having symmetric agents allows us to distinguish between two types of learning phases, selfplay and play with other agents as described in \S \ref{sec:lea}.

\subsection{Learning functions}\label{sec:lea}

In experiment 1, which addresses our first research question on the emergence of the shape bias, agents go through two learning phases: a selfplay phase and then a regular play phase between two distinct agents.\footnote{We motivate this two phase approach in Appendix \ref{app:learn}.} In the initial selfplay phase each agent plays games alone, enacting both the role of speaker and listener. We allow backpropagation of gradients through the message channel from end to end during this phase, since they happen within a single agent. In order to backpropagate through the discrete communication channel, we use the Gumbel-Softmax gradient estimator~\cite{jang2017categorical} similarly to previous work on selfplay \cite{foerster2016learning, havrylov2017emergence, mordatch2018emergence,lowe2020interaction}.

\textbf{Selfplay loss} $\ell_\textsc{sp}$: we use the cross-entropy loss on the final prediction:
\begin{equation} \label{eq:self}
    \ell_\textsc{sp} \coloneqq -\log p_\textsc{sl}(i|i, I)
\end{equation}
where $p_\textsc{sl}(\cdot|i, I)$ is the prediction made by the speaker-listener chain.

Next, agents play together in a community play phase. Since our agents are symmetric, they take turns being either the speaker or the listener; this is randomly assigned with each batch of games. As they are separate agents, we do not allow gradients to pass through the message channel during this phase; instead we rely on reinforcement learning.

\textbf{Community loss} $\ell_{\textsc{c}}$: The community loss is the joint speaker S (\ref{eq:sa}) and listener L (\ref{eq:lc}) losses. S updates using the accuracy reward signal given by $r \coloneqq \mathbb{I}[i=i'] \in \{1,0\}$, where $i'$ is the listener's best guess and $i$ is the ground truth target\footnote{Note that the speaker does not have access to the listener's beliefs in the form of cross-entropy rewards, but instead learns using accuracy rewards, since the speaker should not observe the internal states of the listener.}. The speaker loss $\ell_{\textsc{s}}$ is:
\begin{align}
\ell_{\textsc{s}} &\coloneqq -(r-b) \log p_\textsc{s}(\bm{m}|i) + \ell_\textsc{h} \label{eq:sa}\\
\ell_\textsc{h} &\coloneqq - \frac{c}{|\bm{m}|} \sum_{n=1}^{|\bm{m}|} \sum_{m_n\in\mathcal{V}} p_{\textsc{s}}(m_n) \log \frac{1}{p_{\textsc{s}}(m_n)},
\end{align}
where $b$ is the baseline representing the mean reward across previous batches. We add the speaker's entropy $\ell_\textsc{h}$ as a regularizer, here $c = 0.01$.

The listener learns using cross-entropy. The listener loss $\ell_{\textsc{l}}$ is:
\begin{align}
 \ell_{\textsc{l}} &\coloneqq -\log p_\textsc{l}(i|\bm{m},I) \label{eq:lc}
\end{align}
%The community loss is obtained by summing the speaker and listener losses: $\ell_{\textsc{c}} \coloneqq \ell_{\textsc{s}} + \ell_{\textsc{l}}$.

In experiment 2, which addresses our second research question on the persistence of the shape bias, we use an iterated learning paradigm which is designed to study language evolution across generations of learners \cite{kirby2001spontaneous, kirby2014iterated, libowling2019ease, cogswell2020emergence, Ren2020Compositional, lu2020countering, dagan2020coevolution}. For iterated learning, we introduce an additional student-teacher play phase to those previously described. During this phase, a student -- an agent from the current generation -- and a teacher -- an agent from the previous generation -- play together like regular community play except that the teacher agent does not update, only the student. They take turns being the speaker and the listener. Thus, the loss in student-teacher play either consists of the speaker loss $\ell_{\textsc{s}}$ (\ref{eq:sa}) or the listener loss $\ell_{\textsc{l}}$ (\ref{eq:lc}) depending on the role played by the student.

\section{Measuring Efficiency in Neural Agents}

In order to determine if agents are efficient communicators and whether the shape bias follows the communicative efficiency trade-off, we first define the metrics we will use to estimate communicative need, information loss, and complexity.

There are some differences between our approach from previous metrics that should be noted \cite{kemp2018semantic}. First, we do not assume that communication between a speaker and a listener is always successful; we are in a language learning context where learners make mistakes. For this reason, our informativity measure is based on communicative success rate, rather than the expected cost of needing to refer to an explicit shape. Second, we do not assume that the naming systems these agents learn have perfect coverage. Thus, we cannot define an explicit generative model for lexical categories and count the minimum number of rules necessary to describe a semantic system as a complexity measure, but instead we must rely on estimates of mutual information (MI) between agents' linguistic representations and the available shape classes in our dataset. MI between linguistic representations (words) and semantic classes (meaning) has in fact previously been used to estimate complexity of color naming systems \cite{zaslavsky2019color, Chaabounie2021communicating}.

\textbf{Communicative need} is defined as the proportion of `shape games' played during training. A shape game is a game where the distractors and the target only differ in terms of shape (but share the same color, texture, and size), so it is necessary for the speaker's message to encode information about shape distinction to successfully communicate the target to the listener (The same applies to color games, where the target and distractor objects only differ in terms is color). Agents trained with a higher proportion of shape games have a higher communicative need to distinguish shape classes.

\begin{figure*}[ht]
\vskip 0.2in
\begin{center}
\centerline{\includegraphics[width=\textwidth]{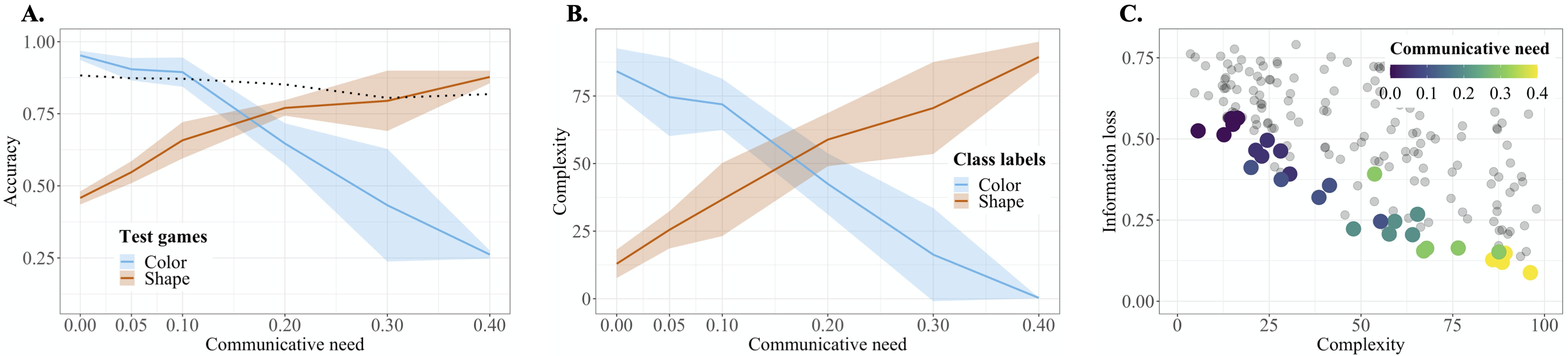}}
\caption{\textbf{A.} Listener accuracy on test games. 0.25 is chance performance. The dotted line represents overall test accuracy. There is a slight drop in overall accuracy as the number of shape games increases since learning to visually recognize and distinguish shape is harder than more localized visual features like color or texture. \textbf{B.} Complexity, or mutual information between the first embedding of messages and class labels. \textbf{C.} Optimal trade-off between information loss and complexity in agents shape distinction systems. Colored points are single runs with different random seeds and communicative need for shape seen during learning. Gray points are random languages generated by permuting the order of agents' language embeddings.
}\label{fig:exp1}
\end{center}
\vskip -0.2in
\end{figure*}
\textbf{Information loss} is defined as the proportion of test games where the listener fails to identify the correct target given the speaker's message, or  $( 1 - \texttt{accuracy} )$. For example, higher accuracy on shape games means smaller information loss about shape in agents' lexical categories.

\textbf{Complexity} is defined as the MI between class labels (the 10 shape and 8 color classes in our dataset) and agents' language embeddings. In order to obtain the distributions over embeddings and class labels, we take the embedding of the first character of a message generated by speakers for all 11,000 test images, as well as the shape/color label for those same images. We chose to use only the first embedding as it is a good baseline for agents' linguistic representations\footnote{We considered MI between more complex representations and class labels; all show the same result pattern as this baseline (see Appendix \ref{app:mi}).}. Given that each embedding contains multiple features (the embedding size is 64), we sum the MI estimates between each feature and the labels to get the estimated MI between the whole embeddings and labels. In this case, we use the method from \citet{ross2014mutual} for estimating MI between a continuous and a discrete distribution. Higher MI indicates more similarity between linguistic representations and a given meaning space. Thus, for example, higher MI between agents' embeddings and test image shape labels indicates a more complex set of shape distinctions in agents' lexical categories.

\section{Experiment 1: The Emergence of a Shape Bias}

Our first research question asks how a bias towards shape can emerge simultaneously in both linguistic and visual representations without conditioning learning on an existing language. We hypothesise that the shape bias emerges due to communicative need and that agents use efficient communication systems when referring to complex referents.

\subsection{Setup}

We have two agents play the signalling games detailed in the previous section, first in selfplay and then in community play. They play 896,000 games total in each phase. Across all runs, the agent and learning hyperparameters are fixed, we only vary our independent variable, communicative need -- the proportion of shape games seen during training (the non-shape games are randomly generated by sampling the target and distractors from a uniform distribution over objects). We measure information loss and complexity using agents' final state after training. For all measures, we report mean and standard deviation across 5 random seed runs.

\subsection{Results and Analysis}

First, let us consider only the lexical categorization system that agents learn for shape. Figure \ref{fig:exp1}.A shows us that accuracy on shape games is modulated by communicative need. As communicative need increases so does accuracy, in other words, information loss decreases. Figure \ref{fig:exp1}.B shows us that complexity is also modulated by communicative need. As the proportion of shape games increases so does MI between shape labels and the first embedding of speaker messages. Figure \ref{fig:exp1}.C shows us that agents develop communication systems for shape that are efficient, trading off between information loss and complexity as a function of communicative need for shape.

Second, now that we have established that agents develop efficient semantic categorization systems for shape, let us consider what happens to other semantic domains, such as color, as we increase the communicative need for shape. In Figure \ref{fig:exp1}.A, we see that as we increase the communicative need for shape, the accuracy on color games decreases, showing the opposite relation to shape game accuracy. In Figure \ref{fig:exp1}.B, we see the same opposite effect where as communicative need for shape increases, MI between color labels and the first embedding of speaker messages decreases. When communicative need is at $0.0$ and all games are randomly sampled, agents have a strong color bias -- higher accuracy on color games than shape games and higher complexity for color labels than shape labels indicate that agents use a convention system that categorizes referent objects based on their color rather than their shape. However, as we increase communicative need for shape, performance on color games and MI between embeddings and color labels both drop and agents develop a strong shape bias.

As we increase communicative need for shape, we are consequently decreasing the need to distinguish color. This happens for two reasons. First, both shape and color are properties of the same set of 320 objects agents are developing a naming convention system for. If they develop a naming system that categorizes objects based on their shape, they can still use this naming system in most randomly generated games and be successful 9/10 times, without ever needing to reference color. Second, as we increase the number of shape games while holding the number of total games fixed, we are consequently decreasing the number of random games and therefore decreasing the number of color games, since about 0.001\% of randomly sampled games are color games.

This suggests that higher communicative need for shape may lead to lower need to communicate other types of distinctions about referent objects, leading to the emergence of a bias towards shape in our lexical categories.

\section{Experiment 2: The Persistence of a Shape Bias} % page 1/2 of 6 and page 7

Our second research question asks whether the existence of a bias towards shape in a parent's lexical categorization system is enough to maintain that bias in new generations of learners. We hypothesise that it is not, and furthermore, propose that a constant external pressure originating from communicative need for shape is necessary in order to maintain a shape bias in lexical categorization systems across generations.

\subsection{Iterated learning}

We use an iterated learning paradigm that allows us to explore how agent languages shift across generations \cite{kirby2001spontaneous, kirby2014iterated, kirby2015compression}. Iterated learning involves having multiple generations of learners where each generation learns from the previous one. These paradigms usually involve some kind of `information bottleneck' that forces language to evolve over time by not allowing perfect noiseless transmission of information between generations.

Different versions of iterated learning have been used with neural agents to consider how structural properties of language can evolve with cultural evolution \cite{cogswell2020emergence, libowling2019ease, Ren2020Compositional, lu2020countering, dagan2020coevolution}. Here, we instead consider how lexical categorization biases can evolve in agents over time.

\subsection{Setup}

The original generation will be composed of the agents with the strongest shape bias from experiment 1: those trained with 40\% shape games. The following generations are trained in one of two condition: with 40\% shape games (high communicative need for shape) or only on random games (low communicative need for shape). This setup will allow us to determine if communicative need must be maintained in order for the shape bias to be maintained as well. At each generation, after an initial selfplay phase, agents play games with a teacher agent sampled from the previous generation before playing with each other in community play. Agents play 896,000 games per phase. We train a total of 14 generations per training condition and report mean and standard deviation across 5 random seed runs for all measures.

\subsection{Results and Analysis}
\begin{figure}[t]
\vskip 0.2in
\begin{center}
\centerline{\includegraphics[width=\columnwidth]{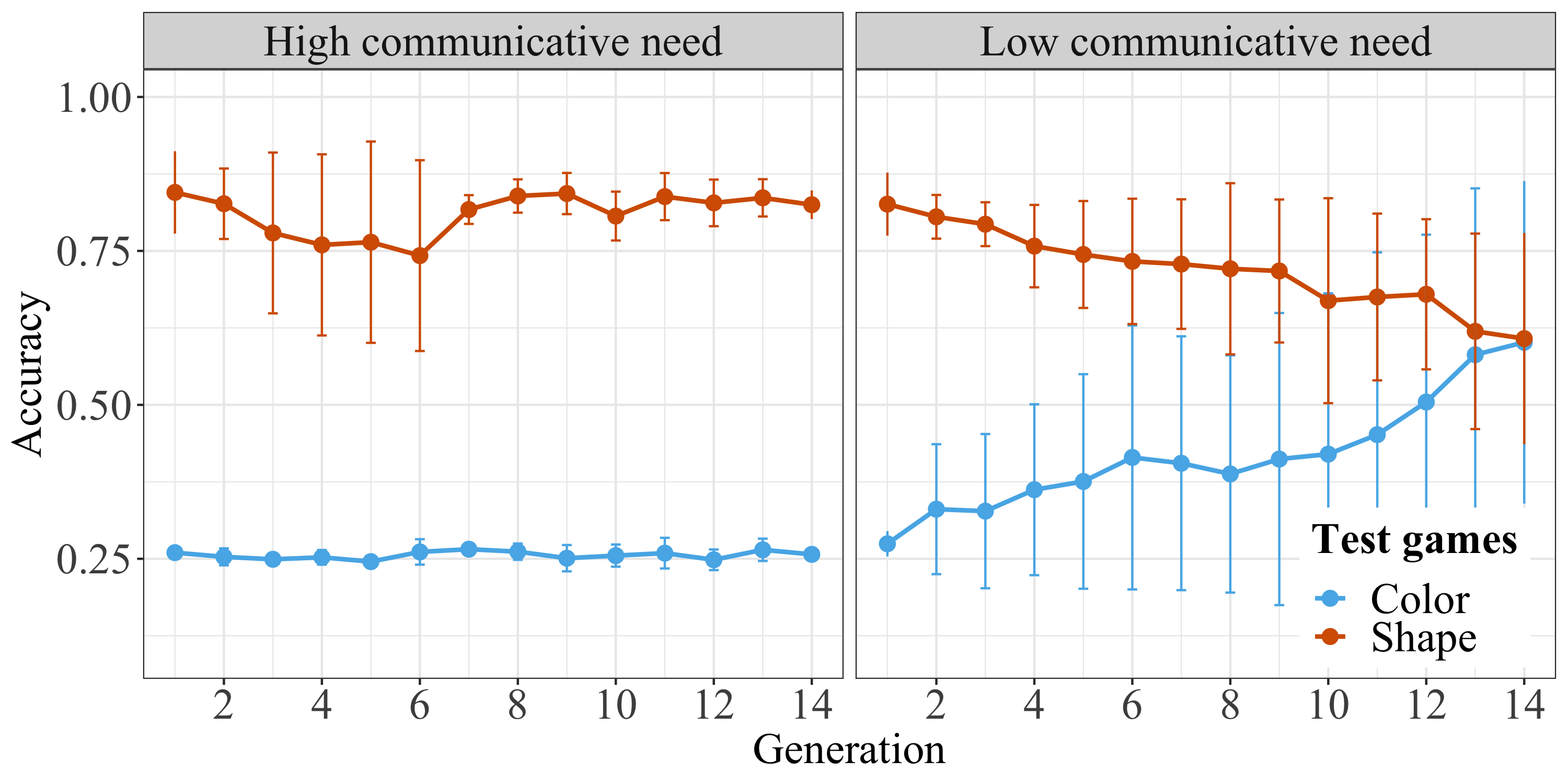}}
\caption{Listener accuracy on shape and color test games at each generation. 0.25 is chance performance.}\label{fig:exp2}
\end{center}
\vskip -0.2in
\end{figure}

In the `high communicative need' condition a bias towards shape is maintained across all generations (see  Figure \ref{fig:exp2}). Here, higher accuracy on test shape games than color games indicates that agents are using a lexical categorization system that makes more shape distinctions than color distinctions. In the `low communicative need' condition, we instead observe that the shape bias disappears after 14 generations and that color game accuracy is trending upward. This suggests that given enough generations agents will eventually return to lexical categorization systems that reflect their \textit{innate} color bias, observed in experiment 1. Unless the shape bias is an innate inductive bias, we expect it to disappear over time if there is little external pressure from communicative need to keep it.

\section{Discussion}

We have shown that a shape bias can emerge and persist across generations of emergent communication agents if there is a high enough communicative need for shape. In this final section, we address what our results may teach us about the shape bias in human learners and neural networks.

Previous work considered the lexicon one semantic domain at a time, but our simulations demonstrate that communicative need affects the structure of the lexicon as a whole. Experiment 1 showed us how the need for distinction in one semantic domain may inadvertently affect another domain and lead to categorization biases, such as the shape bias. Human learners may also experience this need for discriminating shapes in communicative contexts, which could help them learn this bias. Experiment 2 showed that without external pressures we expect that, as language evolves, learners will eventually fall back to using their innate inductive biases to form lexical categories. There is evidence that the shape bias in humans is not innate, but learnt, as it starts around two years old and grows stronger over time \cite{landau1988importance}. These results suggest that the shape bias in human learners may be sustained by communicative need for shape in addition to the learning mechanisms which ensure its transmission from a caregivers' lexical categorizations to children's.

There are several reasons we might expect people to have a high communicative need for shape in the real world. Most importantly, there are functional reasons that motivate the use of shape distinction. In the case of artifacts, shape is often related to an affordance (e.g. `hammer' refers to a class of objects which share the same shape, and that shape is what determines its use). We may often be trying to refer to objects' functions during communication. The shape bias is in fact strongest in contexts where the referents are assumed to be artifacts rather than natural kinds, often even disappearing in the latter context \cite{imai1997cross, soja1991ontological, yoshida2003shifting, diesendruck2003specific, booth2005conceptual, cimpian2005absence}. We would predict this to be the case, since we do not expect a shape bias in contexts with low communicative need for shape.
%we consider two of them. First, ...Second, there may be physiological reasons to believe that shape distinction is important. Shape is for the most part perceived the same way across seeing humans (and even non-seeing humans), while color and texture are not necessarily given that color blindness and myopia are such common conditions.

Other types of neural networks have been shown to develop a shape bias (1) by increasing the size of models or (2) by augmenting their input data \cite{hosseini2018assessing, geirhos2018imagenettrained, hill2019understanding, bhojanapalli2021understanding, tuli2021convolutional}. Note that neither of these approaches can account for the bias being contextually modulated the way it is in humans. When it comes to making models larger (1), we show that it is possible for relatively small models to learn a shape bias, and furthermore, we do not find that the size of agent models affected our experiment 1 results (see Appendix \ref{app:size}). As for using augmented input data (2), this is done by increasing the number of distinct shapes versus other features in a dataset such that objects differ more in terms of shape than other features, for example by having fewer colors, like using black and white images.  The reason these data augmentations increase shape bias is in fact because they increase the need to distinguish shapes during training instances when each training image is seen an even number of times. We show that these artificial data augmentations are not necessary if you increase the model's need to distinguish shape through how it interacts with the data during training (see Appendix \ref{app:shape}). Thus, considering the shape distinction need probability during learning rather than using augmented data may help us develop models with a stronger shape bias while maintaining representative datasets.

In future work, we hope to address the context specificity of the shape bias and consider whether neural agents can learn to have a shape bias in some referential spaces but not others as a function of communicative need across different contexts.

\section*{Acknowledgements}
We would like to acknowledge the contributions of Nicole Fitzgerald and R\'{e}mi Tachet des Combes to earlier iterations of this project. We would also like to thank our anonymous reviews for their helpful comments. The first author would like to thank Steven Rygaard for his valuable technical support and Brandon Waldon for insightful discussions during the pandemic, both of which helped push this project forward. Additionally, we would like to thank the members of the Stanford NLP group, the Cognition and Language Lab, and ALPS Lab for their feedback. The first author is supported by the Social Sciences and Humanities Research Council of Canada (SSHRC) Doctoral Fellowship and the Goodan Family Fellowship.

\bibliography{emerge, proposal, bias} % page 9

\begin{thebibliography}{62}
\expandafter\ifx\csname natexlab\endcsname\relax\def\natexlab#1{#1}\fi

\bibitem[{Andreas and Klein(2016)}]{andreas-klein-2016-reasoning}
Jacob Andreas and Dan Klein. 2016.
\newblock \href {https://www.aclweb.org/anthology/D16-1125} {Reasoning about
  {P}ragmatics with {N}eural {L}isteners and {S}peakers}.
\newblock In \emph{Proceedings of the 2016 Conference on Empirical Methods in
  Natural Language Processing}, pages 1173--1182.

\bibitem[{Bachman et~al.(2019)Bachman, Hjelm, and
  Buchwalter}]{bachman2019amdim}
Philip Bachman, R~Devon Hjelm, and William Buchwalter. 2019.
\newblock \href
  {https://proceedings.neurips.cc/paper/2019/file/ddf354219aac374f1d40b7e760ee5bb7-Paper.pdf}
  {Learning {R}epresentations by {M}aximizing {M}utual {I}nformation {A}cross
  {V}iews}.
\newblock In \emph{Advances in Neural Information Processing Systems},
  volume~32.

\bibitem[{Bhojanapalli et~al.(2021)Bhojanapalli, Chakrabarti, Glasner, Li,
  Unterthiner, and Veit}]{bhojanapalli2021understanding}
Srinadh Bhojanapalli, Ayan Chakrabarti, Daniel Glasner, Daliang Li, Thomas
  Unterthiner, and Andreas Veit. 2021.
\newblock \href {https://arxiv.org/pdf/2103.14586.pdf} {Understanding
  robustness of transformers for image classification}.
\newblock \emph{arXiv preprint arXiv:2103.14586}.

\bibitem[{Booth et~al.(2005)Booth, Waxman, and Huang}]{booth2005conceptual}
Amy~E Booth, Sandra~R Waxman, and Yi~Ting Huang. 2005.
\newblock Conceptual information permeates word learning in infancy.
\newblock \emph{Developmental psychology}, 41:491--505.

\bibitem[{Bouchacourt and Baroni(2018)}]{bouchacourtbaroni2018agents}
Diane Bouchacourt and Marco Baroni. 2018.
\newblock \href {https://www.aclweb.org/anthology/D18-1119} {How agents see
  things: On visual representations in an emergent language game}.
\newblock In \emph{Proceedings of the 2018 Conference on Empirical Methods in
  Natural Language Processing}.

\bibitem[{Bouchacourt and Baroni(2019)}]{bouchacourt-baroni-2019-miss}
Diane Bouchacourt and Marco Baroni. 2019.
\newblock \href {https://www.aclweb.org/anthology/P19-1380} {Miss {T}ools and
  {M}r {F}ruit: Emergent {C}ommunication in {A}gents {L}earning about {O}bject
  {A}ffordances}.
\newblock In \emph{Proceedings of the 57th Annual Meeting of the Association
  for Computational Linguistics}, pages 3909--3918.

\bibitem[{Cao et~al.(2018)Cao, Lazaridou, Lanctot, Leibo, Tuyls, and
  Clark}]{cao2018emergent}
Kris Cao, Angeliki Lazaridou, Marc Lanctot, Joel~Z Leibo, Karl Tuyls, and
  Stephen Clark. 2018.
\newblock \href {https://openreview.net/forum?id=Hk6WhagRW} {Emergent
  {C}ommunication through {N}egotiation}.
\newblock In \emph{International Conference on Learning Representations}.

\bibitem[{Chaabouni et~al.(2021)Chaabouni, Kharitonov, Dupoux, and
  Baroni}]{Chaabounie2021communicating}
Rahma Chaabouni, Eugene Kharitonov, Emmanuel Dupoux, and Marco Baroni. 2021.
\newblock \href {https://doi.org/10.1073/pnas.2016569118} {Communicating
  artificial neural networks develop efficient color-naming systems}.
\newblock \emph{Proceedings of the National Academy of Sciences}, 118.

\bibitem[{Cimpian and Markman(2005)}]{cimpian2005absence}
Andrei Cimpian and Ellen~M Markman. 2005.
\newblock The absence of a shape bias in children's word learning.
\newblock \emph{Developmental Psychology}, 41:1003--1019.

\bibitem[{Cogswell et~al.(2020)Cogswell, Lu, Lee, Parikh, and
  Batra}]{cogswell2020emergence}
Michael Cogswell, Jiasen Lu, Stefan Lee, Devi Parikh, and Dhruv Batra. 2020.
\newblock \href {https://arxiv.org/pdf/1904.09067.pdf} {Emergence of
  compositional language with deep generational transmission}.
\newblock \emph{arXiv preprint arXiv:1904.09067}.

\bibitem[{Dagan et~al.(2020)Dagan, Hupkes, and Bruni}]{dagan2020coevolution}
Gautier Dagan, Dieuwke Hupkes, and Elia Bruni. 2020.
\newblock \href {https://arxiv.org/abs/2001.03361} {Co-evolution of language
  and agents in referential games}.
\newblock \emph{arXiv preprint arXiv:2001.03361}.

\bibitem[{Diesendruck and Bloom(2003)}]{diesendruck2003specific}
Gil Diesendruck and Paul Bloom. 2003.
\newblock How specific is the shape bias?
\newblock \emph{Child development}, 74:168--178.

\bibitem[{Foerster et~al.(2016)Foerster, Assael, de~Freitas, and
  Whiteson}]{foerster2016learning}
Jakob Foerster, Ioannis~Alexandros Assael, Nando de~Freitas, and Shimon
  Whiteson. 2016.
\newblock \href
  {https://proceedings.neurips.cc/paper/2016/file/c7635bfd99248a2cdef8249ef7bfbef4-Paper.pdf}
  {Learning to {C}ommunicate with {D}eep {M}ulti-{A}gent {R}einforcement
  {L}earning}.
\newblock In \emph{Advances in Neural Information Processing Systems},
  volume~29, pages 2137--2145.

\bibitem[{Geirhos et~al.(2019)Geirhos, Rubisch, Michaelis, Bethge, Wichmann,
  and Brendel}]{geirhos2018imagenettrained}
Robert Geirhos, Patricia Rubisch, Claudio Michaelis, Matthias Bethge, Felix~A.
  Wichmann, and Wieland Brendel. 2019.
\newblock \href {https://openreview.net/forum?id=Bygh9j09KX} {Imagenet-trained
  {CNN}s are biased towards texture; increasing shape bias improves accuracy
  and robustness.}
\newblock In \emph{International Conference on Learning Representations}.

\bibitem[{Gibson et~al.(2017)Gibson, Futrell, Jara-Ettinger, Mahowald, Bergen,
  Ratnasingam, Gibson, Piantadosi, and Conway}]{gibson2017color}
Edward Gibson, Richard Futrell, Julian Jara-Ettinger, Kyle Mahowald, Leon
  Bergen, Sivalogeswaran Ratnasingam, Mitchell Gibson, Steven~T Piantadosi, and
  Bevil~R Conway. 2017.
\newblock Color naming across languages reflects color use.
\newblock \emph{Proceedings of the National Academy of Sciences},
  114(40):10785--10790.

\bibitem[{Gibson et~al.(2019)Gibson, Futrell, Piantadosi, Dautriche, Mahowald,
  Bergen, and Levy}]{gibson2019efficiency}
Edward Gibson, Richard Futrell, Steven~P Piantadosi, Isabelle Dautriche, Kyle
  Mahowald, Leon Bergen, and Roger Levy. 2019.
\newblock How efficiency shapes human language.
\newblock \emph{Trends in cognitive sciences}, 23:389--407.

\bibitem[{Harding~Graesser et~al.(2019)Harding~Graesser, Cho, and
  Kiela}]{harding-graesser-etal-2019-emergent}
Laura Harding~Graesser, Kyunghyun Cho, and Douwe Kiela. 2019.
\newblock \href {https://www.aclweb.org/anthology/D19-1384} {Emergent
  {L}inguistic {P}henomena in {M}ulti-{A}gent {C}ommunication {G}ames}.
\newblock In \emph{Proceedings of the 2019 Conference on Empirical Methods in
  Natural Language Processing and the 9th International Joint Conference on
  Natural Language Processing (EMNLP-IJCNLP)}, pages 3700--3710.

\bibitem[{Havrylov and Titov(2017)}]{havrylov2017emergence}
Serhii Havrylov and Ivan Titov. 2017.
\newblock \href
  {https://proceedings.neurips.cc/paper/2017/file/70222949cc0db89ab32c9969754d4758-Paper.pdf}
  {Emergence of {L}anguage with {M}ulti-agent {G}ames: {L}earning to
  {C}ommunicate with {S}equences of {S}ymbols}.
\newblock In \emph{Advances in Neural Information Processing Systems},
  volume~30, pages 2149--2159.

\bibitem[{Heibeck and Markman(1987)}]{heibeck1987word}
Tracy~H Heibeck and Ellen~M Markman. 1987.
\newblock Word learning in children: An examination of fast mapping.
\newblock \emph{Child development}, pages 1021--1034.

\bibitem[{Hill et~al.(2019)Hill, Clark, Hermann, and
  Blunsom}]{hill2019understanding}
Felix Hill, Stephen Clark, Karl~Moritz Hermann, and Phil Blunsom. 2019.
\newblock Understanding early word learning in situated artificial agents.
\newblock \emph{arXiv preprint arXiv:1710.09867}.

\bibitem[{Hosseini et~al.(2018)Hosseini, Xiao, Jaiswal, and
  Poovendran}]{hosseini2018assessing}
Hossein Hosseini, Baicen Xiao, Mayoore Jaiswal, and Radha Poovendran. 2018.
\newblock \href {http://labs.ece.uw.edu/nsl/papers/Assessing_Shape_Bias.pdf}
  {Assessing shape bias property of convolutional neural networks}.
\newblock In \emph{Proceedings of the IEEE Conference on Computer Vision and
  Pattern Recognition Workshops}, pages 1923--1931.

\bibitem[{Imai and Gentner(1997)}]{imai1997cross}
Mutsumi Imai and Dedre Gentner. 1997.
\newblock A cross-linguistic study of early word meaning: Universal ontology
  and linguistic influence.
\newblock \emph{Cognition}, 62:169--200.

\bibitem[{J{\"a}ger(2007)}]{jager2007evolution}
Gerhard J{\"a}ger. 2007.
\newblock The evolution of convex categories.
\newblock \emph{Linguistics and Philosophy}, 30:551--564.

\bibitem[{Jang et~al.(2017)Jang, Gu, and Poole}]{jang2017categorical}
Eric Jang, Shixiang Gu, and Ben Poole. 2017.
\newblock \href {https://openreview.net/forum?id=rkE3y85ee} {Categorical
  reparameterization with {G}umbel-{S}oftmax}.
\newblock \emph{Proceedings of the 5th International Conference on Learning
  Representations, {ICLR}}.

\bibitem[{Johnson et~al.(2017)Johnson, Hariharan, van~der Maaten, Fei-Fei,
  Lawrence~Zitnick, and Girshick}]{johnson2017clevr}
Justin Johnson, Bharath Hariharan, Laurens van~der Maaten, Li~Fei-Fei,
  C~Lawrence~Zitnick, and Ross Girshick. 2017.
\newblock \href {https://arxiv.org/abs/1612.06890} {Clevr: A diagnostic dataset
  for compositional language and elementary visual reasoning}.
\newblock In \emph{Proceedings of the IEEE Conference on Computer Vision and
  Pattern Recognition}, pages 2901--2910.

\bibitem[{Kemp et~al.(2007)Kemp, Perfors, and Tenenbaum}]{kemp2007learning}
Charles Kemp, Amy Perfors, and Joshua~B Tenenbaum. 2007.
\newblock Learning overhypotheses with hierarchical bayesian models.
\newblock \emph{Developmental science}, 10:307--321.

\bibitem[{Kemp and Regier(2012)}]{kemp2012kinship}
Charles Kemp and Terry Regier. 2012.
\newblock Kinship categories across languages reflect general communicative
  principles.
\newblock \emph{Science}, 336:1049--1054.

\bibitem[{Kemp et~al.(2018)Kemp, Xu, and Regier}]{kemp2018semantic}
Charles Kemp, Yang Xu, and Terry Regier. 2018.
\newblock Semantic typology and efficient communication.
\newblock \emph{Annual Review of Linguistics}, 4:109--128.

\bibitem[{Kingma and Ba(2014)}]{kingma2014adam}
Diederik~P Kingma and Jimmy Ba. 2014.
\newblock \href {https://arxiv.org/pdf/1412.6980.pdf} {Adam: A method for
  stochastic optimization}.
\newblock \emph{arXiv preprint arXiv:1412.6980}.

\bibitem[{{Kirby}(2001)}]{kirby2001spontaneous}
Simon {Kirby}. 2001.
\newblock \href {https://ieeexplore.ieee.org/document/918430} {Spontaneous
  evolution of linguistic structure-an iterated learning model of the emergence
  of regularity and irregularity}.
\newblock \emph{IEEE Transactions on Evolutionary Computation}, 5(2):102--110.

\bibitem[{Kirby et~al.(2014)Kirby, Griffiths, and Smith}]{kirby2014iterated}
Simon Kirby, Tom Griffiths, and Kenny Smith. 2014.
\newblock Iterated learning and the evolution of language.
\newblock \emph{Current opinion in neurobiology}, 28:108--114.

\bibitem[{Kirby et~al.(2015)Kirby, Tamariz, Cornish, and
  Smith}]{kirby2015compression}
Simon Kirby, Monica Tamariz, Hannah Cornish, and Kenny Smith. 2015.
\newblock Compression and communication in the cultural evolution of linguistic
  structure.
\newblock \emph{Cognition}, 141:87--102.

\bibitem[{Kottur et~al.(2017)Kottur, Moura, Lee, and
  Batra}]{kottur-etal-2017-natural}
Satwik Kottur, Jos{\'e} Moura, Stefan Lee, and Dhruv Batra. 2017.
\newblock \href {https://www.aclweb.org/anthology/D17-1321} {Natural {L}anguage
  {D}oes {N}ot {E}merge {`}{N}aturally{'} in {M}ulti-{A}gent {D}ialog}.
\newblock In \emph{Proceedings of the 2017 Conference on Empirical Methods in
  Natural Language Processing}, pages 2962--2967.

\bibitem[{Landau et~al.(1988)Landau, Smith, and Jones}]{landau1988importance}
Barbara Landau, Linda~B Smith, and Susan~S Jones. 1988.
\newblock The importance of shape in early lexical learning.
\newblock \emph{Cognitive development}, 3:299--321.

\bibitem[{Lazaridou et~al.(2018)Lazaridou, Hermann, Tuyls, and
  Clark}]{lazaridou2018emergence}
Angeliki Lazaridou, Karl~Moritz Hermann, Karl Tuyls, and Stephen Clark. 2018.
\newblock \href {https://openreview.net/forum?id=HJGv1Z-AW} {Emergence of
  {L}inguistic {C}ommunication from {R}eferential {G}ames with {S}ymbolic and
  {P}ixel {I}nput}.
\newblock In \emph{Proceedings of the 6th International Conference on Learning
  Representations, {ICLR}}.

\bibitem[{Lazaridou et~al.(2017)Lazaridou, Peysakhovich, and
  Baroni}]{lazaridou2017multi}
Angeliki Lazaridou, Alexander Peysakhovich, and Marco Baroni. 2017.
\newblock \href {https://openreview.net/forum?id=Hk8N3Sclg} {Multi-{A}gent
  {C}ooperation and the {E}mergence of ({N}atural) {L}anguage}.
\newblock In \emph{5th International Conference on Learning Representations,
  {ICLR} 2017, Toulon, France, April 24-26, 2017, Conference Track
  Proceedings}.

\bibitem[{Levenshtein(1966)}]{levenshtein1966binary}
Vladimir~I Levenshtein. 1966.
\newblock Binary codes capable of correcting deletions, insertions, and
  reversals.
\newblock In \emph{Soviet physics doklady}, volume~10, pages 707--710.

\bibitem[{Lewis(1969)}]{lewis1969convention}
David Lewis. 1969.
\newblock \emph{Convention: A philosophical study}.
\newblock Harvard University Press, Harvard, Massachusetts.

\bibitem[{Li and Bowling(2019)}]{libowling2019ease}
Fushan Li and Michael Bowling. 2019.
\newblock \href
  {https://proceedings.neurips.cc/paper/2019/file/b0cf188d74589db9b23d5d277238a929-Paper.pdf}
  {{E}ase-of-{T}eaching and {L}anguage {S}tructure from {E}mergent
  {C}ommunication}.
\newblock In \emph{Advances in Neural Information Processing Systems},
  volume~32, pages 15851--15861.

\bibitem[{Lowe et~al.(2020)Lowe, Gupta, Foerster, Kiela, and
  Pineau}]{lowe2020interaction}
Ryan Lowe, Abhinav Gupta, Jakob Foerster, Douwe Kiela, and Joelle Pineau. 2020.
\newblock \href {https://openreview.net/forum?id=rJxGLlBtwH} {On the
  interaction between supervision and self-play in emergent communication}.
\newblock \emph{Proceedings of the 8th International Conference on Learning
  Representations, {ICLR}}.

\bibitem[{Lu et~al.(2020)Lu, Singhal, Strub, Courville, and
  Pietquin}]{lu2020countering}
Yuchen Lu, Soumye Singhal, Florian Strub, Aaron Courville, and Olivier
  Pietquin. 2020.
\newblock \href {http://proceedings.mlr.press/v119/lu20c/lu20c.pdf} {Countering
  {L}anguage {D}rift with {S}eeded {I}terated {L}earning}.
\newblock In \emph{Proceedings of the 37th International Conference on Machine
  Learning}, volume 119, pages 6437--6447.

\bibitem[{Markman(1990)}]{markman1990constraints}
Ellen~M Markman. 1990.
\newblock Constraints children place on word meanings.
\newblock \emph{Cognitive science}, 14(1):57--77.

\bibitem[{Mordatch and Abbeel(2018)}]{mordatch2018emergence}
Igor Mordatch and Pieter Abbeel. 2018.
\newblock \href
  {https://www.aaai.org/ocs/index.php/AAAI/AAAI18/paper/viewFile/17007/15846}
  {Emergence of grounded compositional language in multi-agent populations}.
\newblock In \emph{Proceedings of the AAAI Conference on Artificial
  Intelligence}, volume~32.

\bibitem[{Ohmer et~al.(2020)Ohmer, K{\"o}nig, and
  Franke}]{ohmer2020reinforcement}
Xenia Ohmer, Peter K{\"o}nig, and Michael Franke. 2020.
\newblock \href {https://cogsci.mindmodeling.org/2020/papers/0393/0393.pdf}
  {Reinforcement of semantic representations in pragmatic agents leads to the
  emergence of a mutual exclusivity bias}.
\newblock In \emph{Proceedings of CogSci}, volume~42.

\bibitem[{Ohmer et~al.(2021)Ohmer, Marino, K{\"o}nig, and
  Franke}]{ohmerand2021}
Xenia Ohmer, Michael Marino, Peter K{\"o}nig, and Michael Franke. 2021.
\newblock \href
  {https://www.home.uni-osnabrueck.de/michfranke/Papers/OhmerMarino_2021_Why\%20and\%20how\%20to\%20study\%20the\%20impact\%20of\%20perception\%20on\%20language\%20emergence1.pdf}
  {Why and how to study the impact of perception on language emergence in
  artificial agents}.
\newblock In \emph{Proceedings of CogSci}.

\bibitem[{O’Connor(2014)}]{o2014evolving}
Cailin O’Connor. 2014.
\newblock Evolving perceptual categories.
\newblock \emph{Philosophy of Science}, 81:840--851.

\bibitem[{Quine(1960)}]{quine1960word}
Willard Van~Orman Quine. 1960.
\newblock \emph{Word and object MIT press}.
\newblock MIT Press.

\bibitem[{Regier(2003)}]{regier2003emergent}
Terry Regier. 2003.
\newblock Emergent constraints on word-learning: A computational perspective.
\newblock \emph{Trends in Cognitive Sciences}, 7:263--268.

\bibitem[{Regier(2005)}]{regier2005emergence}
Terry Regier. 2005.
\newblock The emergence of words: {A}ttentional learning in form and meaning.
\newblock \emph{Cognitive science}, 29:819--865.

\bibitem[{Ren et~al.(2020)Ren, Guo, Labeau, Cohen, and
  Kirby}]{Ren2020Compositional}
Yi~Ren, Shangmin Guo, Matthieu Labeau, Shay~B. Cohen, and Simon Kirby. 2020.
\newblock \href {https://openreview.net/forum?id=HkePNpVKPB} {Compositional
  languages emerge in a neural iterated learning model}.
\newblock In \emph{International Conference on Learning Representations}.

\bibitem[{Ritter et~al.(2017)Ritter, Barrett, Santoro, and
  Botvinick}]{ritter2017cognitive}
Samuel Ritter, David~GT Barrett, Adam Santoro, and Matt~M Botvinick. 2017.
\newblock \href {https://arxiv.org/abs/1706.08606} {Cognitive psychology for
  deep neural networks: A shape bias case study}.
\newblock In \emph{International conference on machine learning}, pages
  2940--2949.

\bibitem[{Ross(2014)}]{ross2014mutual}
Brian~C Ross. 2014.
\newblock Mutual information between discrete and continuous data sets.
\newblock \emph{PloS one}, 9:e87357.

\bibitem[{Samuelson(2002)}]{samuelson2002statistical}
Larissa~K Samuelson. 2002.
\newblock Statistical regularities in vocabulary guide language acquisition in
  connectionist models and 15-20-month-olds.
\newblock \emph{Developmental psychology}, 38:1016 -- 1037.

\bibitem[{Samuelson and Smith(1999)}]{samuelson1999early}
Larissa~K Samuelson and Linda~B Smith. 1999.
\newblock Early noun vocabularies: do ontology, category structure and syntax
  correspond?
\newblock \emph{Cognition}, 73:1--33.

\bibitem[{Simion and Giorgio(2015)}]{simion2015face}
Francesca Simion and Elisa~Di Giorgio. 2015.
\newblock \href {https://www.ncbi.nlm.nih.gov/pmc/articles/PMC4496551/} {Face
  perception and processing in early infancy: inborn predispositions and
  developmental changes}.
\newblock \emph{Frontiers in psychology}.

\bibitem[{Smith and Samuelson(2006)}]{smith2006attentional}
Linda~B Smith and Larissa Samuelson. 2006.
\newblock An attentional learning account of the shape bias: Reply to cimpian
  and markman (2005) and booth, waxman, and huang (2005).
\newblock \emph{Developmental psychology}, 42:1339--1343.

\bibitem[{Soja et~al.(1991)Soja, Carey, and Spelke}]{soja1991ontological}
Nancy~N Soja, Susan Carey, and Elizabeth~S Spelke. 1991.
\newblock Ontological categories guide young children's inductions of word
  meaning: Object terms and substance terms.
\newblock \emph{Cognition}, 38:179--211.

\bibitem[{Steels and Belpaeme(2005)}]{steels2005coordinating}
Luc Steels and Tony Belpaeme. 2005.
\newblock Coordinating perceptually grounded categories through language: A
  case study for colour.
\newblock \emph{Behavioral and Brain Sciences}, 28:469–--529.

\bibitem[{Tuli et~al.(2021)Tuli, Dasgupta, Grant, and
  Griffiths}]{tuli2021convolutional}
Shikhar Tuli, Ishita Dasgupta, Erin Grant, and Thomas~L Griffiths. 2021.
\newblock \href {https://arxiv.org/pdf/2105.07197.pdf} {Are convolutional
  neural networks or transformers more like human vision?}
\newblock \emph{arXiv preprint arXiv:2105.07197}.

\bibitem[{Xu et~al.(2020)Xu, Liu, and Regier}]{xu2014numeral}
Yang Xu, Emmy Liu, and Terry Regier. 2020.
\newblock \href {https://doi.org/10.1162/opmi_a_00034} {Numeral systems across
  languages support efficient communication: From approximate numerosity to
  recursion}.
\newblock In \emph{Open mind : discoveries in cognitive science}, volume~4,
  pages 57--70.

\bibitem[{Yoshida and Smith(2003)}]{yoshida2003shifting}
Hanako Yoshida and Linda~B Smith. 2003.
\newblock Shifting ontological boundaries: how japanese-and english-speaking
  children generalize names for animals and artifacts.
\newblock \emph{Developmental Science}, 6:1--17.

\bibitem[{Zaslavsky et~al.(2019)Zaslavsky, Kemp, Tishby, and
  Regier}]{zaslavsky2019color}
Noga Zaslavsky, Charles Kemp, Naftali Tishby, and Terry Regier. 2019.
\newblock Color naming reflects both perceptual structure and communicative
  need.
\newblock \emph{Topics in cognitive science}, 11:207--219.

\end{thebibliography}
\bibliographystyle{acl_natbib}

\clearpage

\appendix

\section{Exemplar Images from Dataset} \label{app:data}

\begin{figure}[H]
%\vskip 0.2in
\begin{center}
\centerline{\includegraphics[scale=0.6]{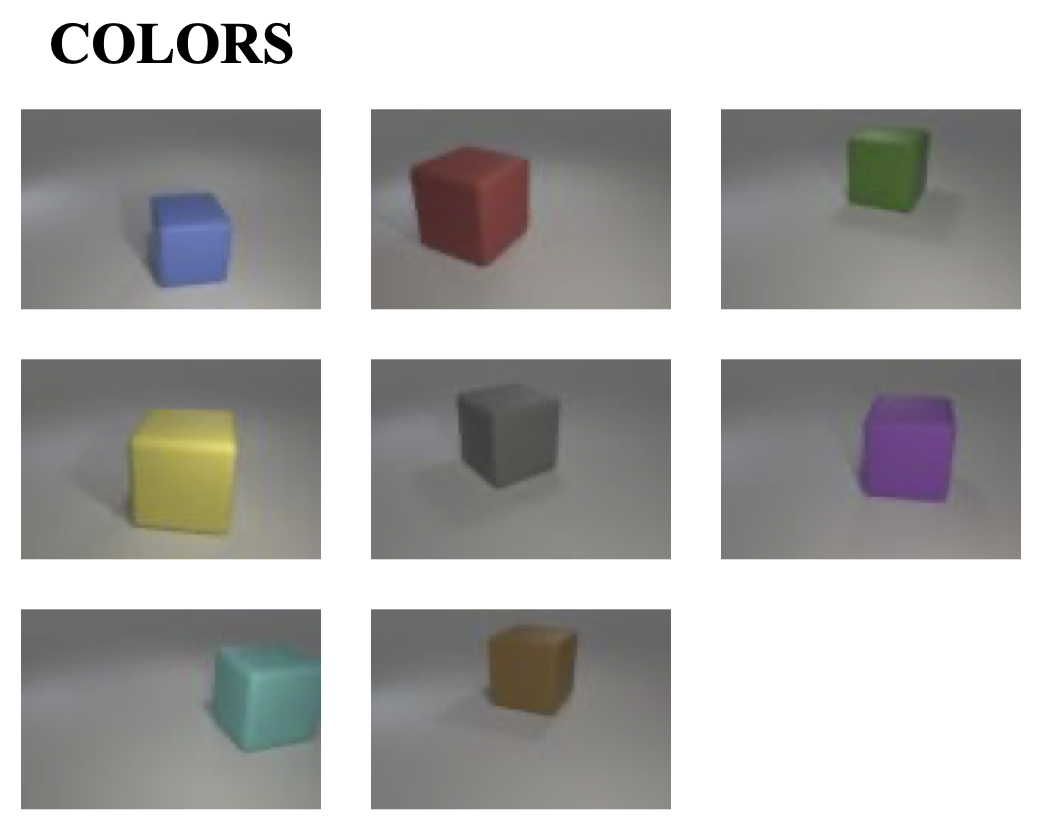}}
\caption{All colors: blue, red, green, yellow, gray, purple, cyan, brown. }
\end{center}
%\vskip -0.2in
\end{figure}

\begin{figure}[H]
%\vskip 0.2in
\begin{center}
\centerline{\includegraphics[scale=0.6]{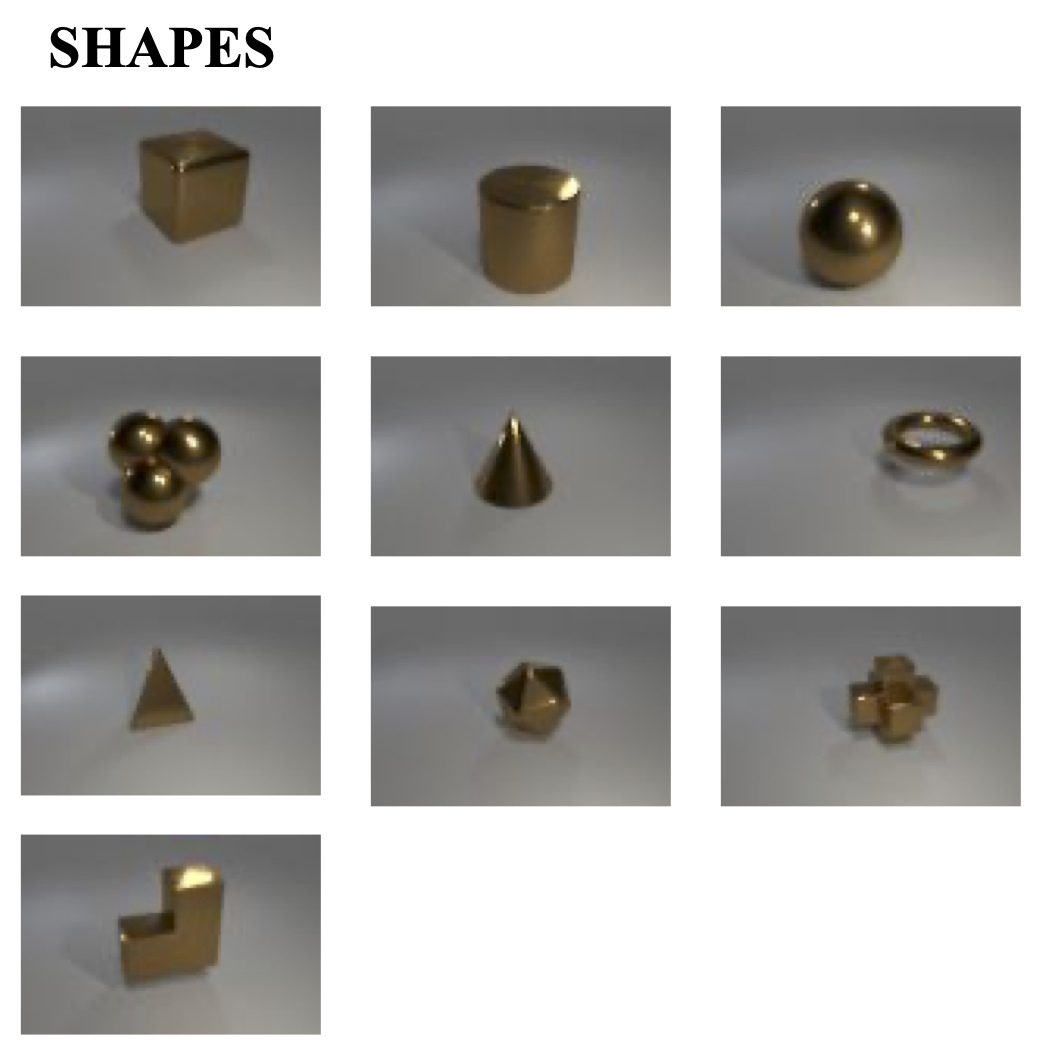}}
\caption{All shapes: cube, cylinder, sphere, tetrahedral sphere form, cone, torus, tetrahedron, icosahedron, 6-sided symmetric cross, 3D L-shape.}
\end{center}
%\vskip -0.2in
\end{figure}

\begin{figure}[H]
\vskip 0.2in
\begin{center}
\centerline{\includegraphics[scale=0.6]{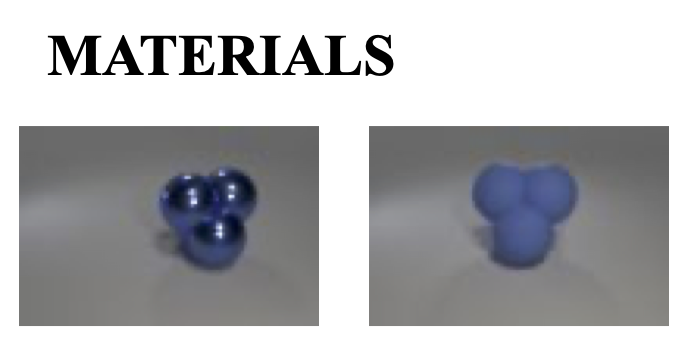}}
\caption{All materials: metal, rubber.}
\end{center}
\vskip -0.2in
\end{figure}

\begin{figure}[H]
\vskip 0.2in
\begin{center}
\centerline{\includegraphics[scale=0.6]{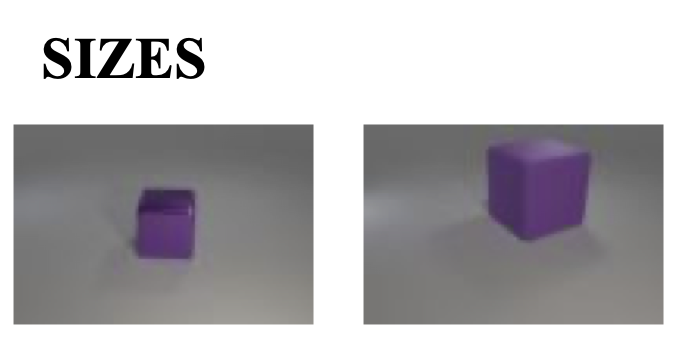}}
\caption{All sizes: small, large.}
\end{center}
\vskip -0.2in
\end{figure}

\section{Hyperparameters}\label{app:hyp}
Here are the hyperparameters tested, bolded ones are the settings we used for the experiments reported in the main paper.
\paragraph{Game hyperparameters: \\}
\textit{number of distractor}s: \textbf{3}
\\\textit{number of games per epoch}: 6,400 / 32,000 / \textbf{64,000} / 96,000

\paragraph{Learning hyperparameters: \\}
\textit{number of epochs per phase}: 10 / 12 / \textbf{14} / 15  \\
\textit{learning rate}: \textbf{0.001} / 0.0001 / 0.00001 \\
\textit{batch size}: 32 / 128 / 512 / \textbf{1024} / 2048

\paragraph{Population hyperparameters: \\}
\textit{population size}: \textbf{2} \\
\textit{number of pairs}: \textbf{1} (since population size is 2, only pair (0,1) is possible)\\
\textit{number of generations}: \textbf{0} (for experiment 1); 7, \textbf{14} (for experiment 2) \\
\textit{random seed}: \textbf{[1:5]}

\paragraph{Agent vision hyperparameters: \\}
\textit{image size}: \textbf{128} (for image resizing) \\
\textit{compression size}: 256, \textbf{512}, 768

\paragraph{Agent language hyperparameters: \\}
\textit{hidden size}: 128, \textbf{256}, 512 \\
\textit{hidden mlp size}: 128, \textbf{256}, 512 \\
\textit{embedding size}: 64 \\
\textit{message length}: 7 \\
\textit{vocabulary size}: 60

\section{Model Specification}

\paragraph{Perception module:} AMDIM model cloned from \url{https://github.com/Philip-Bachman/amdim-public} and trained on our whole dataset for 500 epochs with 200 images per batch. We consider the `rkhs\_1' output layer of size 1024. This part of the model does not update during learning.

\paragraph{Representation module:} An encoder composed of: a linear transformation from size 1024 (AMDIM default output size) to \textit{compression size}, a non-linear Tanh transformation, a linear transformation from \textit{compression size} to \textit{hidden size}.

\paragraph{Production module:} An LSTM Cell with input \textit{embedding size} and output \textit{hidden size}; A linear transformation layer from \textit{hidden size} to \textit{vocabulary size}. We consider the output at each state for \textit{message length} states.

\paragraph{Comprehension module:} An LSTM with input \textit{embedding size} and output \textit{hidden size}; An MLP consisting of a linear transformation from \textit{hidden size} to \textit{hidden mlp size}, a non-linear ReLU transformation, a linear transformation from \textit{hidden mlp size} to \textit{hidden size}. We consider the final state after \textit{message length} states.

\paragraph{Message classifier:} This is not part of the agents, but the classifier we used to evaluate \textit{complexity} of different agent languages. It is composed of a single linear layer of input size message length and of output size number of shape(10)/color(8) categories.

\paragraph{Simple CNN vision module:} This module is not contained in our regular agents but used in our experiment in appendix C where we compare our vision setup to this simple CNN setup. This module can replace the perception and representation modules defined above. It is composed of: A Conv2D layer with 3 input channels, 20 output channels, a kernel size 5 and stride of 1; A ReLU non linear transformation; A max pool 2D layer with kernel size 2 and stride of 2; Another Conv2D layer with 20 input channels, 50 output, kernel of 5 and stride of 1; A ReLU non linear transformation; A max pool 2D layer with kernel size 2 and stride of 2; A batchnorm2D layer; a two layered MLP with dropout at 0.7 that reduces the output to \textit{hidden size}.

\paragraph{Optimizer:} We used an Adam optimizer across all models during training \cite{kingma2014adam}.

\section{Model and Learning Procedure Development} \label{app:learn}
The developmental experiments reported in this section were completed using an earlier version of our dataset which only contained 10,000 images that varied in shape (3 possible shape), color (8), material (2), and size (2). The following experiments were also conducted using slightly different hyperparameters from those used in the main paper.

\subsection{Vision Module Comparison}

\begin{figure}[t]
\vskip 0.2in
\begin{center}
\centerline{\includegraphics[width=\columnwidth]{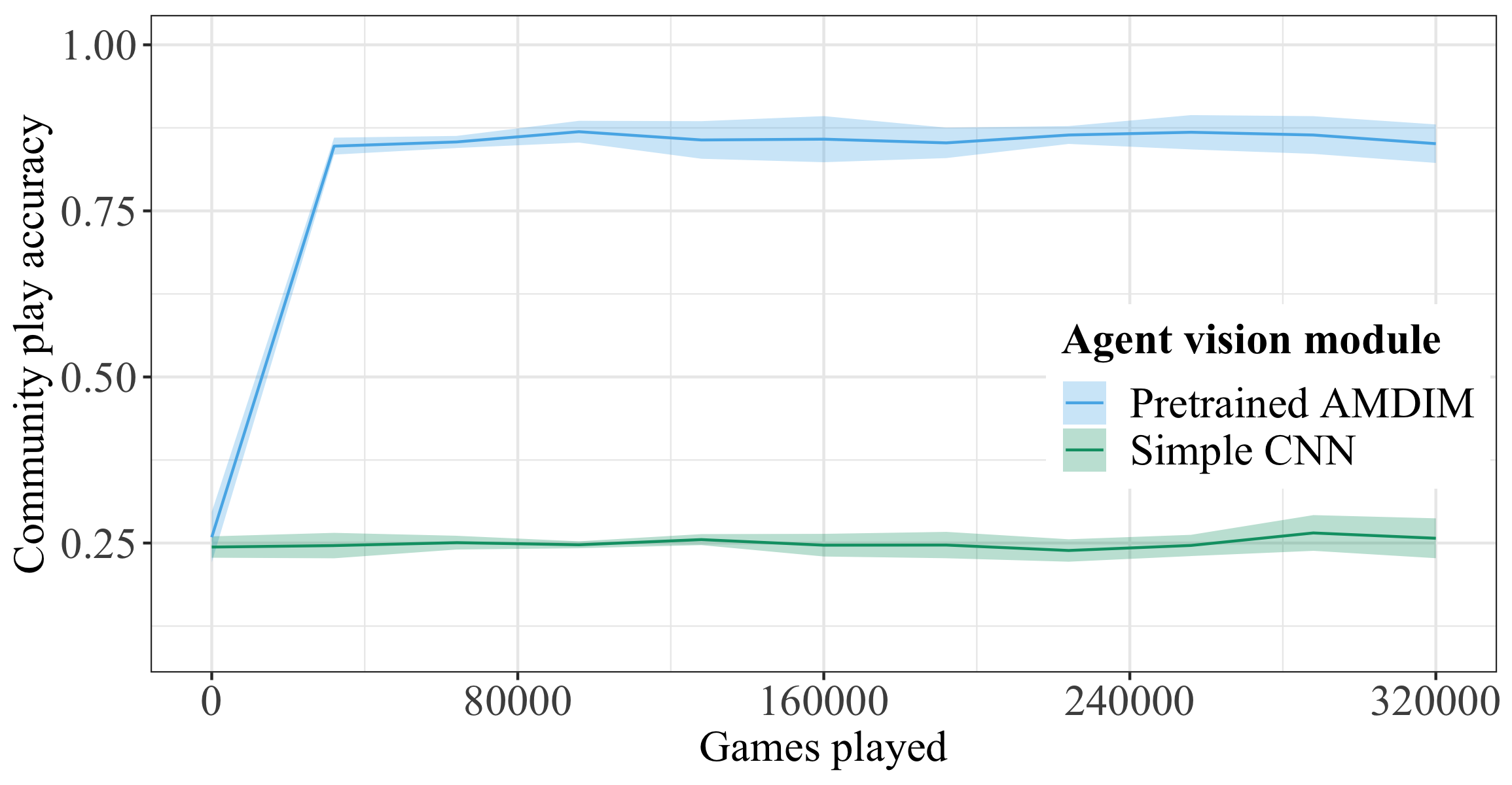}}
\caption{Mean accuracy on test across 5 runs with different random seeds. Lines represent mean and ribbons represent standard deviation. Both models trained using 10,000 steps of 32 selfplay games followed by 10,000 steps of 32 community play games. \label{fig:3}}
\end{center}
\vskip -0.2in
\end{figure}

We compare our vision module approach to previous ones which involved training a simple multi-layered CNN type architecture that processes the images while also learning to communicate about them \cite{lazaridou2018emergence}. This previous approach was notoriously difficult to train and could not guarantee that agents would manage to converge on any communication system at all. We find that using our perception/representation split with pretrained AMDIM allows us to guarantee quick convergence on a successful language, while we could not manage to converge on any successful communication system with the simple CNN setup in so few games, see Figure \ref{fig:3}.

\subsection{Learning functions comparison}

We explore the effects of different learning functions and play phases on agents' capacity to learn to communicate successfully to find effective and efficient learning approaches. Within a single generation, we consider two types of play: regular game play between two agents -- called \textit{community play} (\S \ref{sub:comm}) -- and a form of single-agent play -- \textit{selfplay} (\S \ref{sub:self}). In iterated learning across multiple generations, we consider different forms of student-teacher interactions, aka \textit{intergenerational play} phases (\S \ref{sub:teach}), in addition to regular intragenerational community play. For each of these phases -- community play, selfplay, intergenerational play -- we define a set of possible learning functions.

\subsection{Community play} \label{sub:comm}
We test two different learning functions in \S~\ref{sub:gen0}.

\paragraph{Accuracy loss $\ell_{\textsc{acc}}$:} The first loss jointly updates the speaker S in (\ref{eqa:sa}) and the listener L in (\ref{eqa:la}) using the accuracy reward signal given by $r \coloneqq \mathbb{I}[i=i'] \in \{1,0\}$, where $i'$ is the listener's best guess and $i$ is the ground truth target. This strategy is standard and has previously been used in both the single generation and iterated learning contexts~\citep{lazaridou2018emergence, libowling2019ease}. We start with the speaker-accuracy loss $\ell_{\textsc{sa}}$:
\begin{align}
\ell_{\textsc{sa}} &\coloneqq -(r-b) \log p_\textsc{s}(\bm{m}|i) + \ell_\textsc{h} \label{eqa:sa}\\
\ell_\textsc{h} &\coloneqq - \frac{c}{|\bm{m}|} \sum_{n=1}^{|\bm{m}|} \sum_{m_n\in\mathcal{V}} p_{\textsc{s}}(m_n) \log \frac{1}{p_{\textsc{s}}(m_n)},
\end{align}
where $b$ is the baseline representing the mean reward across previous batches. We add the speaker's entropy as a regularizer: $\ell_\textsc{h}$, where $c$ is a coefficient between 0.1 and 0.01 that varies based on the reward $c = 0.1 - 0.1|r - b|$. We follow up with the listener-accuracy loss $\ell_{\textsc{la}}$:
\begin{align}
\ell_{\textsc{la}} &\coloneqq -(r-b)\log p_\textsc{l}(i'|\bm{m},I) \label{eqa:la}
\end{align}

The accuracy loss is obtained by summing the speaker-accuracy and listener-accuracy losses: $\ell_{\textsc{acc}} \coloneqq \ell_{\textsc{sa}} + \ell_{\textsc{la}}$.

\paragraph{Cross-entropy loss $\ell_{\textsc{ce}}$:} The second approach allows the listener to learn using cross-entropy (\ref{eqa:lc}),  while still using the accuracy-reward based updates for the speaker (\ref{eqa:sa}). We now define the listener-cross-entropy loss $\ell_{\textsc{lc}}$:
\begin{align}
 \ell_{\textsc{lc}} &\coloneqq -\log p_\textsc{l}(i|\bm{m},I) \label{eqa:lc}
\end{align}

Cross-entropy loss is the sum of the speaker-accuracy loss and the listener-cross-entropy loss: $\ell_{\textsc{ce}} \coloneqq \ell_{\textsc{sa}} + \ell_{\textsc{lc}}$.

\subsection{Selfplay} \label{sub:self}
Unlike in community play, selfplay only involves a single agent that plays both the role of the speaker and the listener. We allow backpropagation of gradients through the message channel from end to end. In order to backpropagate through the discrete communication channel, we use the Gumbel-Softmax gradient estimator~\cite{jang2017categorical}.

\textbf{Selfplay loss} $\ell_\textsc{sp}$: we simply use the cross-entropy loss on the final prediction:
\begin{equation} \label{eqa:self}
    \ell_\textsc{sp} \coloneqq -\log p_\textsc{sl}(i|i, I)
\end{equation}
where $p_\textsc{sl}(\cdot|i, I)$ is the prediction made by the speaker-listener chain.

\subsection{Intergenerational play}
We compare three types of student-teacher learning interactions.
\label{sub:teach}

\paragraph{Student-teacher play:}
Student-teacher play is like community play except one of the agents is from the previous generation, while the other is from the current one. The teacher agent does not update, only the student can learn. They take turns being the speaker and the listener. Thus, the loss in student-teacher play either consists of the speaker-accuracy loss $\ell_{\textsc{sa}}$ (\ref{eqa:sa}) or the listener-cross-entropy loss $\ell_{\textsc{lc}}$ (\ref{eqa:lc}) depending on the role played by the student.

\paragraph{Imitation-selfplay loss $\ell_{\textsc{isp}}$:}
Here we combine selfplay with teacher message imitation. We introduce a reward if the student produces a message that is similar to what a teacher would produce given the input image. This similarity reward is given by taking the normalized Levenshtein distance \cite{levenshtein1966binary} between the student message $\bm{m}^s$ and the teacher message $\bm{m}^t$.
\begin{equation}\label{eqa:simreward}
    \sigma(\bm{m}^s,\bm{m}^t) \coloneqq 1 - \frac{\operatorname{lev}(\bm{m}^s,\bm{m}^t)}{|\bm{m}^s|}
\end{equation}

We use this similarity reward to motivate the student to speak in a fashion similar to the teacher. This leads to the student-teacher-similarity loss $\ell_\textsc{sts}$:
\begin{align}
\ell_\textsc{sts} &\coloneqq -(\sigma(\bm{m}^s,\bm{m}^t)-b)\log p_\textsc{s}(\bm{m}^s |i) + \ell_\textsc{h}\label{eqa:ss}.
\end{align}
The student-teacher-similarity loss is added to the selfplay loss (\ref{eqa:self}) to build the imitation-selfplay loss: $\ell_{\textsc{isp}} \coloneqq \ell_\textsc{sts} + \ell_\textsc{sp}$.

\paragraph{Imitation-KD loss $\ell_{\textsc{ikd}}$:}
Knowledge distillation is a form of selfplay where the student tries to learn from its mistakes using the teacher's feedback. Instead of trying to fit its predictions to the ground-truth label of the game $i$, the student tries to fit them to the predicted choices of the teacher given the student's messages, $i^t$, where $i^t \sim (p^t_\textsc{l}(\cdot|\bm{m}^s, I))$ . Like selfplay, this method involves a single student agent playing both the speaker and the listener and is therefore differentiable end to end. The knowledge-distillation loss $\ell_{\textsc{kd}}$ is the cross-entropy loss with $i^t$ as our target image:
\begin{equation} \label{eqa:kd}
    \ell_{\textsc{kd}} \coloneqq -\log p_\textsc{sl}(i^t|i, I)
\end{equation}
The student imitates the teacher's listening capabilities but also has to imitate its speaking abilities, and thus, the imitation-KD loss $\ell_{\textsc{ikd}}$ combines both the  knowledge-distillation loss and the student-teacher-similarity loss (\ref{eqa:ss}): $\ell_{\textsc{ikd}} \coloneqq \ell_\textsc{sts} + \ell_\textsc{kd}$.

% 2 Pages
\subsection{Training generation zero} \label{sub:gen0}
Our goal in this experiment is to compare the two learning functions defined for community play in \ref{sub:comm}, the accuracy loss $\ell_{\textsc{acc}}$ and the cross-entropy loss $\ell_{\textsc{cd}}$, in addition to the effect of introducing an initial selfplay phase prior to community play. We compare four different learning strategies: community play with the accuracy loss, community play with the cross-entropy loss, and then each of these again with an initial phase of selfplay. Within each setup the population, or pair of agents, is trained for 20,000 steps, each with 32 games. In the selfplay case, the first 10,000 steps are dedicated to selfplay and the rest to community play.

\begin{figure}[t]
\vskip 0.2in
\begin{center}
\centerline{\includegraphics[width=\columnwidth]{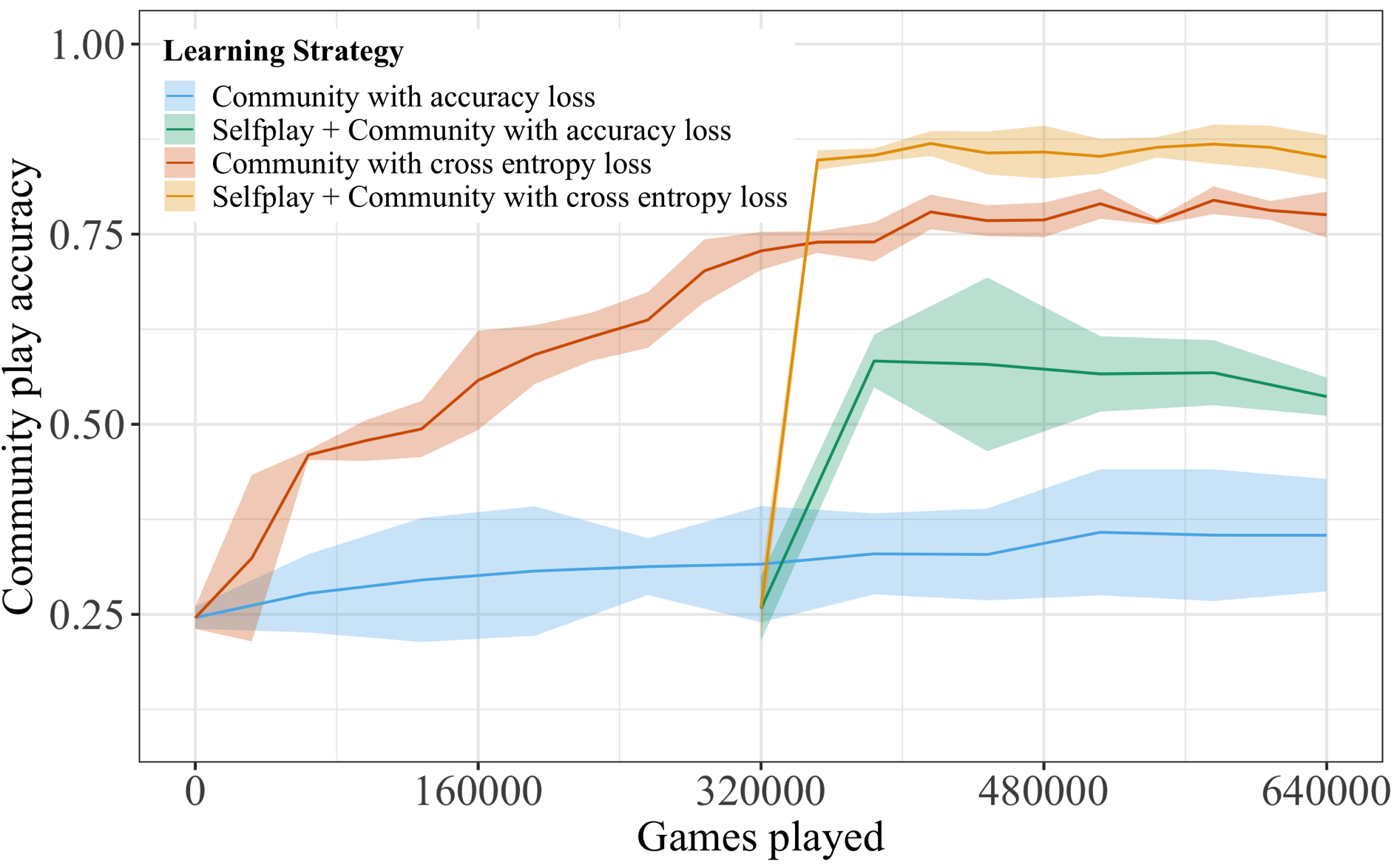}}
\caption{Accuracy on the test set over the number of games played during training. Lines represent mean across 5 random seed runs and the ribbons are standard deviation. Chance performance is 0.25. We compare agents with and without a selfplay phase before community play using accuracy loss $\ell_{\textsc{acc}}$ or cross-entropy loss $\ell_{\textsc{ce}}$. \label{fig:2}}
\end{center}
\vskip -0.2in
\end{figure}

\begin{figure}[ht]
\vskip 0.2in
\begin{center}
\centerline{\includegraphics[width=\columnwidth]{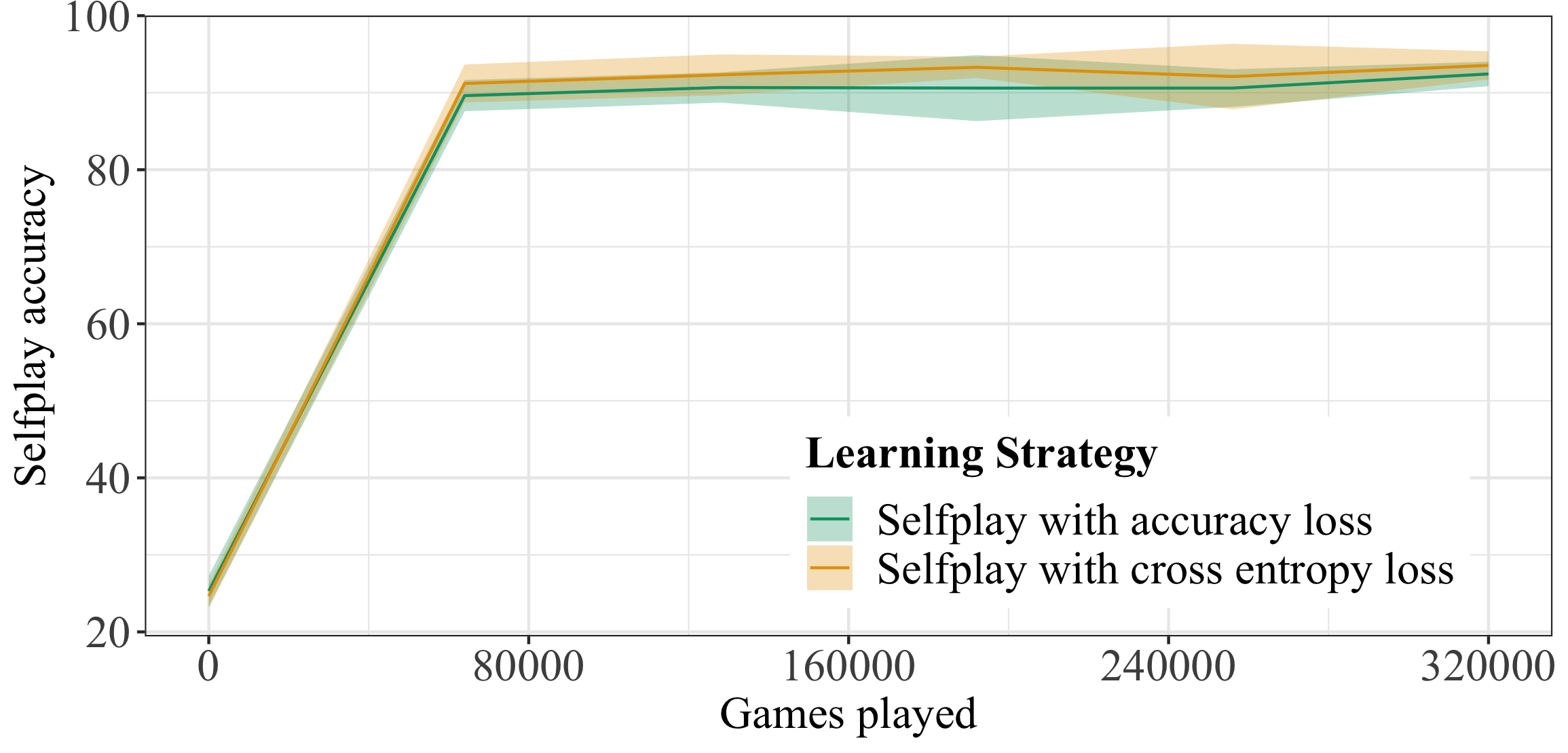}}
\caption{Accuracy on the test set for the first 320,000 games of selfplay using the selfplay loss $\ell_{\textsc{sp}}$.  \label{fig:4}}
\end{center}
\vskip -0.2in
\end{figure}

\paragraph{Results:} We ran each learning strategy 5 times with different random seeds. We report mean performance and the standard deviation across runs. Figure \ref{fig:2} illustrates game accuracy on the test set of unseen combinations in community play based on the number of games played during training. Models with selfplay start halfway as they have spent the first 320,000 games in a selfplay phase (Figure \ref{fig:4}). We find that using the cross-entropy loss during community play leads to much higher accuracy and less variance between runs than using the accuracy loss. Additionally we find that having an initial phase of selfplay ultimately leads to more effective communication during community play.

\subsection{Iterated learning} \label{sub:iter}

In iterated learning we consider learning across multiple generation. In this experiment we test the effect of the different learning functions defined in \ref{sub:teach} during \textit{intergenerational play} on eventual intragenerational \textit{community play}. We consider three different intergenerational learning strategies: (1) student-teacher play phase using the cross-entropy loss $\ell_{\textsc{ce}}$ ; (2) imitation-selfplay using the imitation-selfplay loss $\ell_{\textsc{isp}}$ ; (3) knowledge distillation using the imitation-KD loss $\ell_{\textsc{ikd}}$. We follow each intergenerational play phase by intragenerational community play where we use cross-entropy loss $\ell_{\textsc{ce}}$. The initial generation of teachers for all models is trained using an initial selfplay phase followed by the community play phase. At each subsequent generation, student agents are randomly initialized while the teacher agent is randomly sampled from the previous generation. We train each model for a total of 8 generations (generation zero + 7 student-teacher generations). Each generation was trained for an initial 2,000 steps of 32 games of their respective form of intergenerational play followed by 2,000 steps of community play.

\begin{figure}[ht]
\vskip 0.2in
\begin{center}
\centerline{\includegraphics[width=\columnwidth]{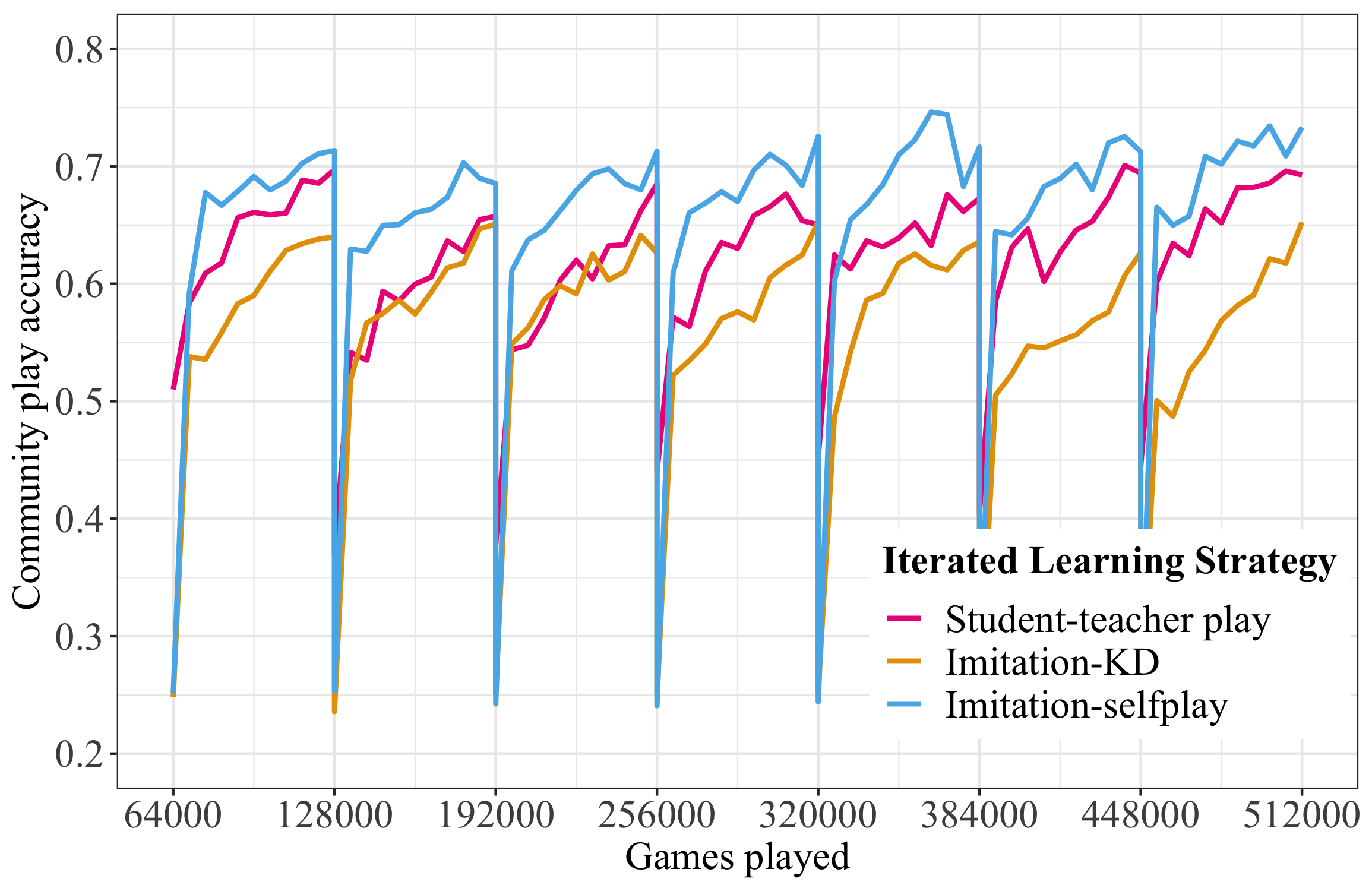}}
\caption{Mean accuracy on test during community play across all seven generation. Chance performance is 0.25. Standard deviation is not reported here for clarity.\label{fig:5}}
\end{center}
\vskip -0.2in
\end{figure}

\begin{figure}[h]
\vskip 0.2in
\begin{center}
\centerline{\includegraphics[width=\columnwidth]{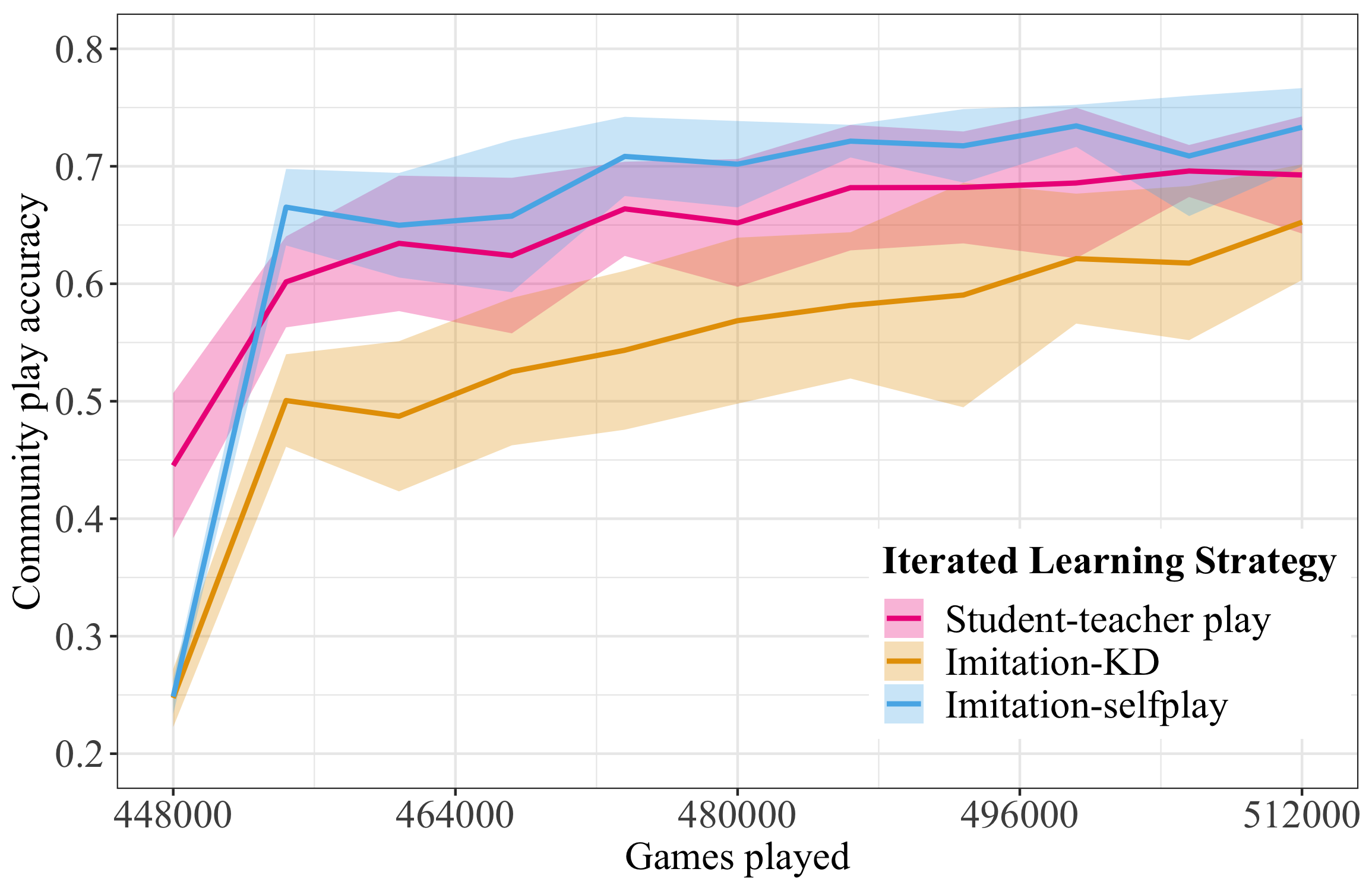}}
\caption{Accuracy on test during generation 7's community play. \label{fig:6}}
\end{center}
\vskip -0.2in
\end{figure}

\paragraph{Results:}
We ran each setup 5 times with different random seeds. We report the mean performance and standard deviation across runs. Figure \ref{fig:5} shows mean accuracy on community play across all generations, while Figure \ref{fig:6} zooms in on generation 7. In Figure \ref{fig:5} compares each intergenerational learning strategy's effect on the subsequent phase of community play at each generation. we see that, as expected, performance drops at each new generation before climbing back up. Between each of these generation `drops', we train the models for a phase of intergenerational learning using their respective learning function. The most effective learning approach seems to be imitation-selfplay loss $\ell_{\textsc{isp}}$. However, a closer look at Figure \ref{fig:6} shows that it is also worth considering simply using student-teacher play since there is no significant difference between imitation-selfplay and student-teacher play. Furthermore, we find that agents trained with student-teacher play have an initial advantage when they start their community play phase, starting above chance, around 0.45.

\section{Mutual Information Estimates} \label{app:mi}
We want to measure mutual information (MI) between linguistic representations and meanings as an estimate of complexity. It is clear that we can consider the color/shape label of the images in our dataset as the intended meanings, however it was less clear what level of linguistic representation we should consider.  We decided to test multiple: (1) the embedding of the first character generated by a speaker for each image, (2) the predicted class of a linear classifier trained on the embedding of the first character generated by a speaker for each image, (3) the whole message generated by a speaker for each image (using the character with highest probability at each step), and (4) the predicted class of a linear classifier trained on the whole message generated by a speaker for each image.
\begin{figure}[H]
\vskip 0.2in
\begin{center}
\centerline{\includegraphics[scale=0.085]{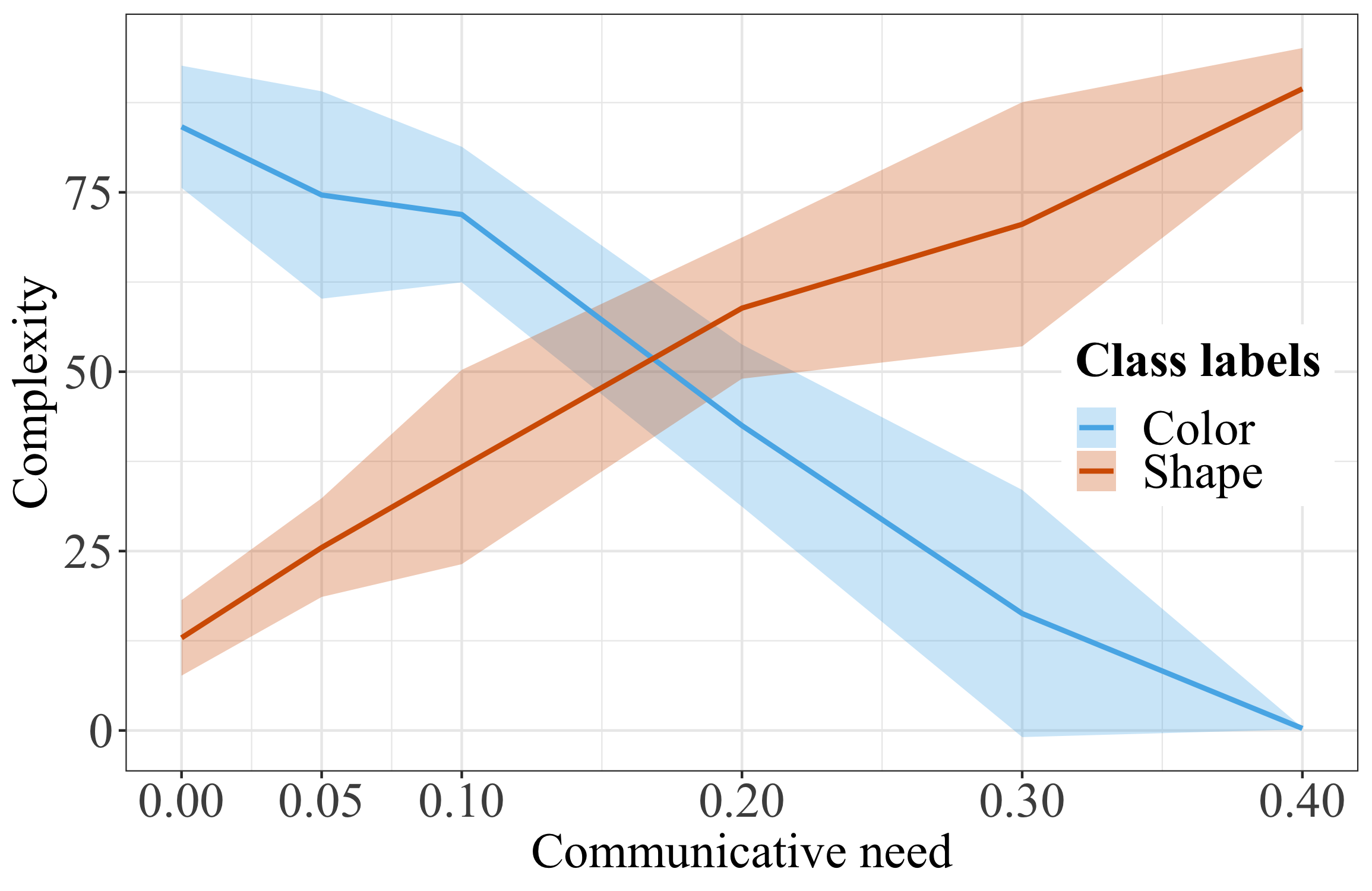}}
\caption{MI between (1) first character embeddings and labels.}
\end{center}
\vskip -0.2in
\end{figure}
\begin{figure}[H]
\vskip 0.2in
\begin{center}
\centerline{\includegraphics[scale=0.085]{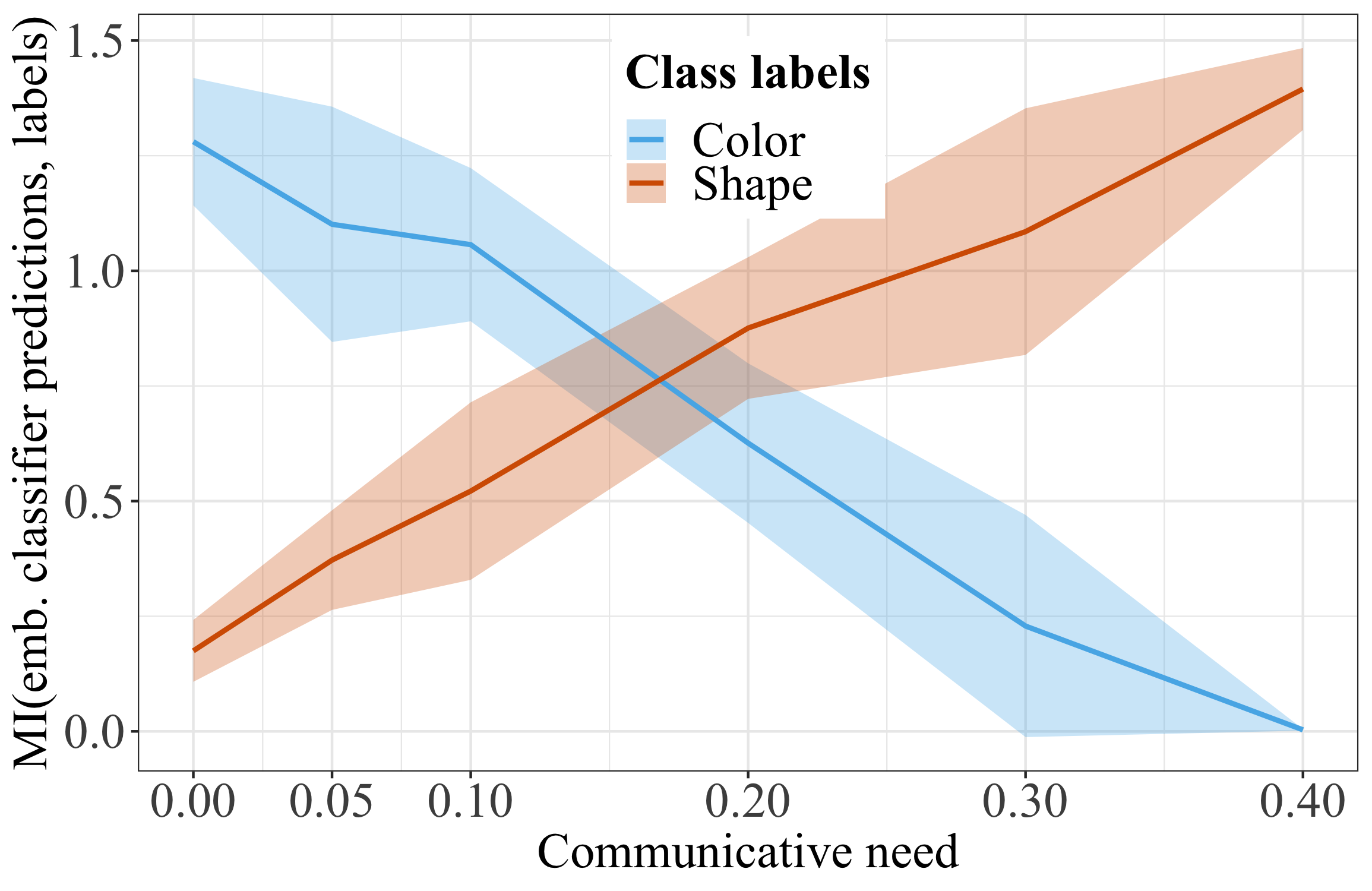}}
\caption{MI between (2) predictions of probe trained on first character embeddings and labels.}
\end{center}
\vskip -0.2in
\end{figure}
\begin{figure}[H]
\vskip 0.2in
\begin{center}
\centerline{\includegraphics[scale=0.085]{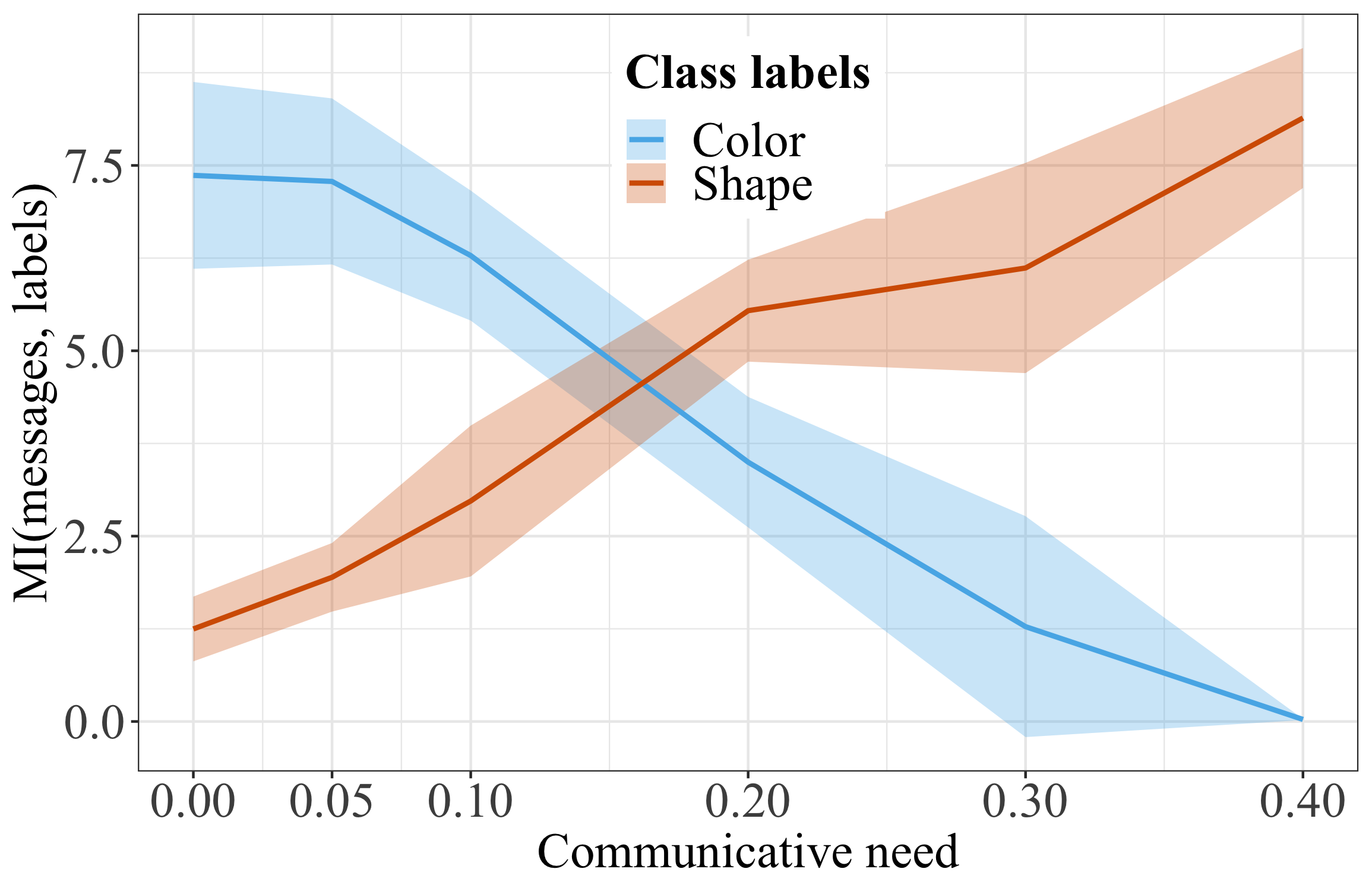}}
\caption{MI between (3) whole messages and labels.}
\end{center}
\vskip -0.2in
\end{figure}
\begin{figure}[H]
\vskip 0.2in
\begin{center}
\centerline{\includegraphics[scale=0.085]{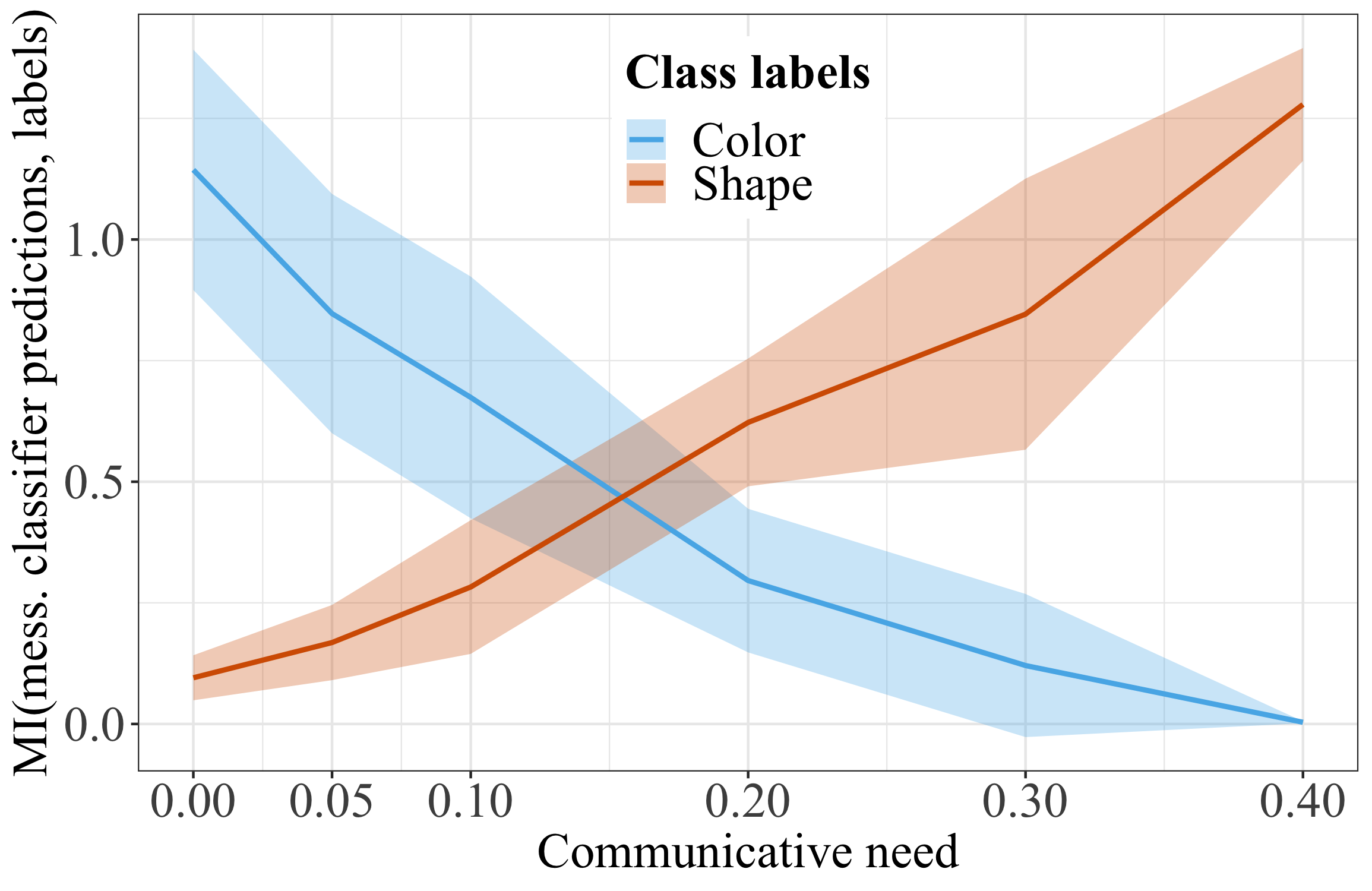}}
\caption{MI between (4) predictions of probe trained on whole messages and labels.}
\end{center}
\vskip -0.2in
\end{figure}

All of these linguistic representations had the same mutual information result pattern, so we chose to report the simplest of these measures in the main paper.

\section{Experiment 1B: More Shapes}\label{app:shape}
\begin{figure}[H]
\vskip 0.2in
\begin{center}
\centerline{\includegraphics[width=\columnwidth]{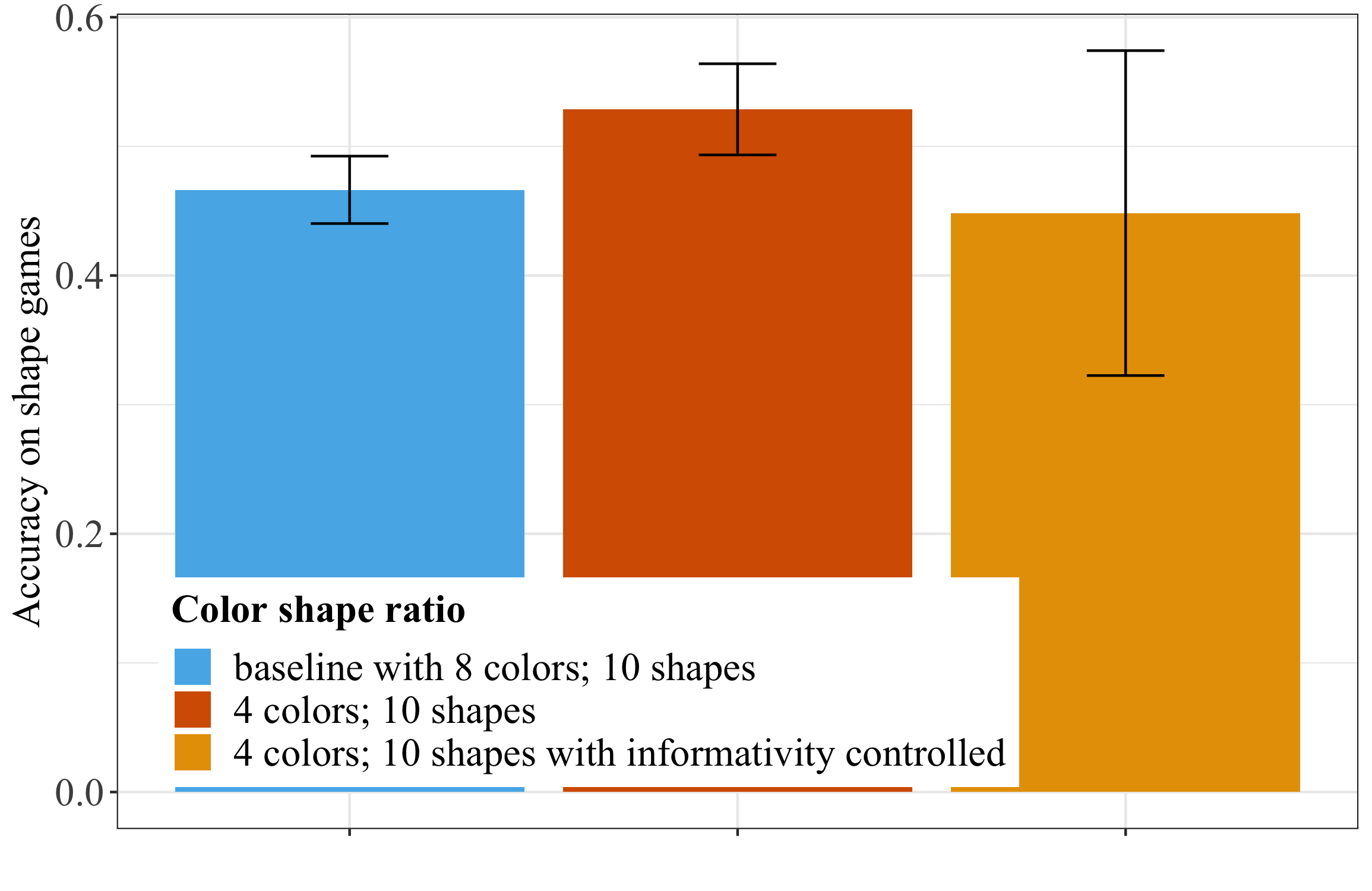}}
\caption{Accuracy on test shape games for agents trained on games sampled from a dataset containing either 4 or 8 distinct color categories. Both the left most and center models were trained on randomly sample games from a uniform distribution over images. The right most played color games 10\% of the time.}\label{fig:exp1B}
\end{center}
\vskip -0.2in
\end{figure}

We test whether increasing the difference in number of shapes versus colors overall in the image dataset leads to higher shape game accuracy (see Figure \ref{fig:exp1B}). We find that it does. We check whether this advantage disappears if we control for shape informativity by increasing the communicative need for color to 10\% to counterbalance the increased number of shapes. We find that the increase in shape game performance disappears, suggesting that data augmentations, in the form of more shape classes, to induce shape bias, lead to a bias not because of increased number, but because the increased number of shape classes leads to higher need to distinguish shape categories. We ran 3 random seed runs for each and report mean and standard deviation.

\section{Experiment 1C: Model Size}\label{app:size}
\begin{figure}[h]
\vskip 0.2in
\begin{center}
\centerline{\includegraphics[scale=0.08]{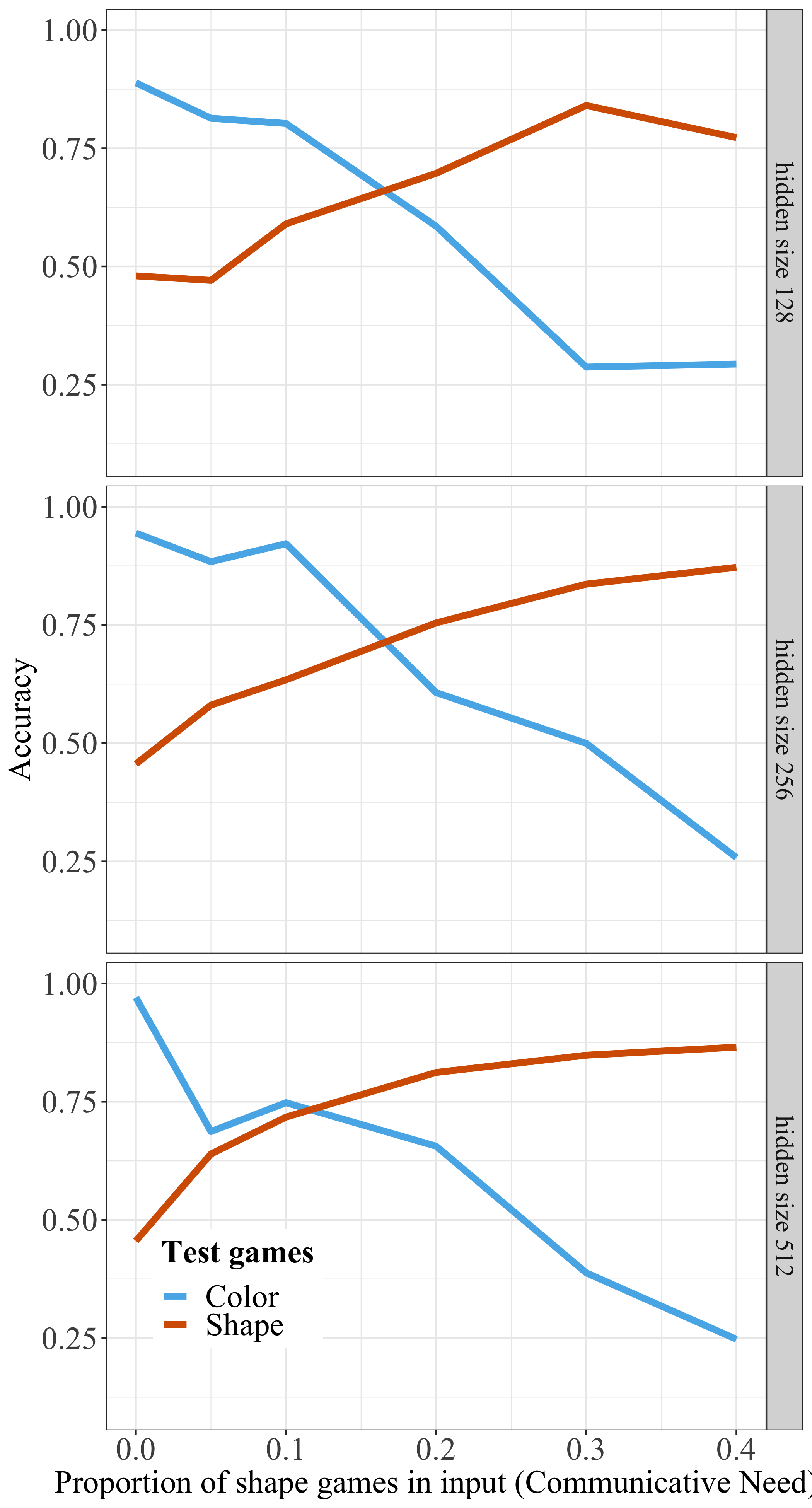}}
\caption{Accuracy on test games for 3 different model sizes as a function of communicative need.}\label{fig:exp1C}
\end{center}
\vskip -0.2in
\end{figure}

We compare three different agent model sizes (see Figure \ref{fig:exp1C}. The middle size, with 256 as its hidden size, is the model size used in the main paper experiments. The smaller model, with hidden size 128, is half the overall size of the middle one, an the larger model, with hidden size 512, is about twice the size of the middle model. We ran 2 random seeds for each and report the mean. We found that the emergence of a shape bias was modulated by communicative need for shape in all cases, regardless of the model size.

\end{document}